\documentclass[acmlarge,nonacm]{acmart}

\AtBeginDocument{%
  }

\usepackage{multirow}
\usepackage{makecell}
\usepackage{pifont}
\newcommand{\cmark}{\ding{51}}%
\newcommand{\xmark}{\ding{55}}%

\begin{document}

\title{HARMES: A Multi-Modal Dataset for Wearable Human Activity Recognition with Motion, Environmental Sensing and Sound}


\author{Robin Burchard}
\email{robin.burchard@uni-siegen.de}
\orcid{0000-0002-4130-5287}
\affiliation{%
  \institution{University of Siegen}
  \streetaddress{Hölderlinstr. 3A}
  \city{Siegen}
  \country{Germany}
  \postcode{57076}
}

\author{Pascal-André Brückner}
\email{pascal.brueckner.1992@gmx.de}
\orcid{0009-0000-5249-6202}
\affiliation{%
  \institution{University of Siegen}
  \city{Siegen}
  \country{Germany}
}

\author{Marius Bock}
\email{bock@iai.uni-bonn.de}
\orcid{0000-0001-7401-928X}
\affiliation{%
  \institution{University of Bonn}
  \institution{\& Lamarr Institute for Machine Learning and Artificial Intelligence}
  \city{Bonn}
  \country{Germany}
}
\author{Juergen Gall}
\email{gall@iai.uni-bonn.de}
\orcid{0000-0002-9447-3399}
\affiliation{%
  \institution{University of Bonn}
  \institution{\& Lamarr Institute for Machine Learning and Artificial Intelligence}
  \city{Bonn}
  \country{Germany}
}

\author{Kristof Van Laerhoven}
\email{kvl@eti.uni-siegen.de}
\orcid{0000-0001-5296-5347}
\affiliation{%
  \institution{University of Siegen}
  \city{Siegen}
  \country{Germany}
}

\renewcommand{\shortauthors}{Burchard et al.}


\begin{abstract}
With each sensing modality exhibiting inherent strengths and limitations, multi-modal approaches for wearable Human Activity Recognition (HAR) are becoming increasingly relevant -- particularly for recognizing Activities of Daily Living (ADLs), where individual modalities often produce ambiguous signals for similar or complex activities. This work introduces HARMES, a multi-modal wearable dataset combining three wrist-recorded modalities: motion sensing via an Inertial Measurement Unit (IMU), atmospheric environmental sensors (humidity, temperature, and pressure), and audio. Collected from 20 participants performing household activities in their own homes, HARMES totals over 80 hours of recorded data, with approximately three hours of labeled activity data per participant across 15 ADL classes. To the best of our knowledge, HARMES is the first dataset to combine this particular sensor trio, and it is nearly six times larger than the previously largest wrist-inertial-acoustic HAR dataset. In an extensive benchmark, we evaluate cross-subject generalization and conduct an ablation study revealing that modality contributions are activity-dependent and can provide complementary value, particularly for activities that are ambiguous from motion data alone. HARMES is freely available at Zenodo\footnote{\url{https://doi.org/10.5281/zenodo.19425718}}, alongside example code for loading the dataset and training models on GitHub\footnote{\url{https://github.com/RBurchard/HARMES}}.
\end{abstract}





\maketitle

\section{Introduction}
Human Activity Recognition (HAR) is a research field with many applications in the real-world, such as healthcare \cite{wangSurveyWearableSensor2019,deMultimodalWearableSensing2015,cristinaAudioVideoBasedHuman2024}, assistance for the elderly \cite{caoLeveragingWearablesAssisting2023}, fitness or sports tracking \cite{bockWEAROutdoorSports2024,hoelzemannHangTimeHARBenchmark2023,zhuangSportRelatedHumanActivity2019}, behavioral tracking and compliance monitoring \cite{zhangSmartRingWearable2019}, and many more. While single-modality approaches using, e.g., only IMU(s) or only microphones have been shown to be sufficient for recognizing simple activities \cite{jungHumanActivityClassification2020b,ordonezDeepConvolutionalLSTM2016,nordenAutomaticDetectionSubjective2022}, each modality exhibits inherent strengths and limitations, with a tendency to become ambiguous for more complex or similar activities. Oftentimes, it is described that single modalities can be confounded by activities that are similar with regard to the measured modality. For instance, hand washing and brushing teeth have similar repetitive motion patterns and are thus hard to distinguish using IMUs alone \cite{wahlRealtimeDetectionObsessivecompulsive2023}, while it is impossible even for human annotators to decide which vegetable is being washed with audio recordings alone \cite{huhEpicSoundsLargescaleDataset2025}. Consequently, multi-modal HAR has emerged as a key strategy, consistently improving performance over uni-modal systems \cite{niSurveyMultimodalWearable2024,munznerCNNbasedSensorFusion2017a,guoWearableSensorBased2016}. This is particularly important in real-world applications such as healthcare, assisted living, and smart environments, where reliable recognition of Activities of Daily Living (ADLs) is required. At the same time, effectively integrating heterogeneous modalities remains a challenging research problem, underscoring the need for comprehensive multi-modal datasets for development and evaluation.

As state-of-the-art multi-modal datasets for HAR are becoming larger and larger, default combinations which are included in such datasets are RGB(d)-video data with or without audio \cite{damenRescalingEgocentricVision2022}, and IMU recordings from wearable sensors worn in different positions on the body, often head, wrists, and/or ankles \cite{bockWEAROutdoorSports2024}. While recent research has shown the complementarity of these sensors for HAR \cite{bockWEAROutdoorSports2024}, the real-world applicability of such sensor setups is limited, as using cameras in public and private spaces, e.g., in bathrooms, restaurants, or universities, comes with legal constraints and ethical concerns. In contrast to the context that can be gathered and learned from videos, recent works have shown that affordable atmospheric environmental sensors, such as humidity, pressure, and temperature sensors, can be utilized to support the contextualization of activities that are otherwise ambiguous to be recognized from IMU data alone \cite{barnaStudyHumanActivity2019,burchardImprovedStrategiesMultimodal2026,karandeRaisingBarOmeterIdentifying2025}. Similarly, audio \cite{minnenRecognizingDiscoveringHuman2005,yeonWatchHARRealtimeOndevice2025,mollynSAMoSASensingActivities2022,hayashiDailyActivityRecognition2015,garcia-cejaMultiviewStackingActivity2018,crucianiFeatureLearningHuman2020a,lotfiComparisonAudioIMU2020} and contextual information, such as location (GPS) or Bluetooth beacons \cite{samyounIWashSmartwatchHandwashing2021,kirstenExploringWearableBasedDetection2025}, are also often used for multi-modal HAR, mostly in combination with IMUs. Although audio data inherently also suffers from privacy issues, research has partially circumvented these concerns by degrading the audio signal \cite{liangCharacterizingEffectAudio2020}, calculating lightweight representations \cite{iravantchiPrivacyMicUtilizingInaudible2021}, or on-device model predictions \cite{yeonWatchHARRealtimeOndevice2025}. 

With the affordability, usability, and reduced impact on users' privacy of atmospheric and audio-based sensors having been proven to be effective for HAR, as of today, no publicly available dataset jointly captures wrist-worn IMU data, audio recordings, and atmospheric environmental sensing. Additionally, existing wrist-inertial-acoustic HAR datasets are limited in size, with the largest one capturing 14.2\,h of activity data \cite{mollynSAMoSASensingActivities2022}.
We thus present \textit{HARMES}, a benchmark dataset of motion, environmental, and sound sensor recordings consisting of more than 80 hours of data of participants performing indoor household and kitchen activities of daily living, recorded in the participants' homes. Totaling 20 participants, we provide around three hours of 141 separate repetitions of 15 labeled activities of daily living (ADLs) per participant, as well as one additional hour of various other free-form and partially labeled activities per participant, recorded in diverse environments. To the best of our knowledge, this makes \textit{HARMES} the largest labeled HAR dataset combining wrist-recorded IMU and audio data, while additionally combining these modalities with the rarely explored atmospheric sensors.  
With \textit{HARMES}, we aim to allow the HAR community to explore the following research question: To what extent do environmental and acoustic sensing modalities complement inertial measurements for recognizing activities of daily living, particularly those that are ambiguous from motion data alone?

Our contributions are threefold:
\begin{enumerate}
    \item We introduce HARMES, a large-scale, labeled multi-modal dataset for wearable Human Activity Recognition, combining wrist-worn IMU data, audio recordings, and atmospheric environmental sensing (humidity, temperature, and pressure), thereby covering a modality combination not present in existing public datasets.
    \item We provide an extensive benchmark analysis, including data quality analysis, synchronization across modalities, and an evaluation protocol for cross-subject generalization showing that HARMES is sufficiently large to train state-of-the-art deep learning models that can generalize to previously unseen subjects for person-specific, unscripted styles of activity execution.
    \item In an effort to address our underlying research question, we investigate the contribution of each sensing modality through an ablation study, demonstrating how environmental and audio signals complement inertial measurements, particularly for activities that are ambiguous from motion data alone.
\end{enumerate}

The dataset, code, and detailed access instructions are available in the Meta Appendix (Section \ref{asec:meta}). Please see Section \ref{assec:data_code_avail} for full details on how to obtain and use our code and data.

\section{Related Work}

\subsection{Inertial-Acoustic Activity Recognition}
Previous work has explored fusion models with IMU and microphone data for HAR. For example, Ward et al. used arm and wrist IMUs combined with microphone recordings to detect various workshop-related activities \cite{wardActivityRecognitionAssembly2006}. 
Liang et al. \cite{liangAudioIMUEnhancingInertial2022} use a student-teacher approach to recognize 23 activity classes in a study with 15 participants, in which a model trained on IMU and audio data is used as a teacher to an IMU-only model, leading to an augmented IMU model with higher classification accuracy than a model trained on IMU data alone. In a work by Bhattacharya et al. \cite{bhattacharyaLeveragingSoundWrist2022}, wrist IMU and audio data from the dominant wrist are employed to classify 23 activities in one semi-naturalistic and one in-the-wild evaluation. They show that in-the-wild activity recognition remains challenging, despite achieving up to 94\,\% F1 scores on semi-naturalistic data, and that fine-tuning on novel participants partially mitigates these challenges. Zhuang et al. \cite{zhuangDetectingHandHygienic2023} apply the combination of single-wrist motion and sound data to hand-hygienic behaviours in a study with eight participants over around 50 minutes each and achieve strong classification performances on the task of detecting and classifying such behaviours between daily behaviours. Yang et al. \cite{yangFusionInertialHighResolution2025a} put an emphasis on the use of privacy-preserving acoustic sensing and employ high-resolution audio in the ultrasonic range for capturing environmental information, along with wrist IMU data for HAR.  
Siddiqui et al. and Becker et al. \cite{siddiquiMultimodalHandGesture,beckerGestEarCombiningAudio2019}
show that hand-gesture recognition also benefits from the fusion of wrist-IMU and audio data, achieving user-independent F1 scores of over 97\,\%.

In a study by Arakawa et al. \cite{arakawaPrISMTrackerFrameworkMultimodal2022}, both IMU and wrist audio data are used in a framework for multimodal procedure tracking. The authors utilize sound and motion data to show that augmenting a HAR system with a graph-based procedure representation and user querying leads to an improved procedure step tracking performance.

Overall, prior work clearly demonstrates that combining inertial and acoustic data significantly and reliably improves HAR  and gesture recognition performance and warrants further exploration. 

\subsection{Available Inertial-, Atmospheric and Audio-based Datasets}
To the best of the authors' knowledge, no dataset exists that includes wrist-worn IMU data together with audio recordings and atmospheric sensing. However, previous works exist on audio-augmented inertial activity recognition and the combination of inertial data with environmental sensing. We list relevant datasets that provide a combination of IMU data with one of our two other modalities in Table \ref{tab:datasets_new}. We limit the dataset comparisons to datasets that supply at least one of wrist-IMU and audio data.


\begin{table} 
\centering
\caption{List of available multi-modal IMU datasets, for comparison with our novel HARMES dataset. We include HAR datasets that supply IMU data together with additional modalities. We compare the number (\#Classes) and type of activity classes (Type, S~=~Sports, D~=~Daily Living, C~=~Cooking/Kitchen, F~=~Fall), number of participants (\#Part.), total duration (\#Hours) of labeled data and whether data was recorded at multiple location (Multi-E.). Additionally, we list whether wrist IMUs are present (LW~=~ left wrist, RW~=~right wrist), and whether audio data and environmental atmospheric sensors (Env) are available. $^*$Subset which provides both IMU and activity annotations $^{**}$Exact participant counts are not provided; labels summarize key-steps $\dag$ Activity verb classes. $\ddag$ Excludes calibration periods.}
\label{tab:datasets_new}
\begin{tabular}{llllclccccc}
\multicolumn{1}{c}{\multirow{2}{*}{Dataset}} & \multicolumn{4}{c}{General}                   & \multicolumn{1}{c}{} & \multicolumn{4}{c}{Modalities}             \\ \cline{2-5} \cline{7-10} 
\multicolumn{1}{c}{}                         & \#Part.          & \#Classes  & \#Hours & Multi-E.      & Type                 & LW     & RW     & Audio  & Env   \\ \midrule
CSL-SHARE \cite{liuCSLSHAREMultimodalWearable2021}  & 20               & 22         & \textless{}11 & \xmark & D,S                  & \xmark & \xmark & \cmark & \xmark  \\
WEAR \cite{bockWEAROutdoorSports2024}  & 22               & 18         & $\sim$19      & \cmark & S                    & \cmark & \cmark & \xmark & \xmark  \\
CMU-MMAC \cite{delatorrefernandoGuideCarnegieMellon2008}        & 34$^{**}$        & 16$^\dag$  & \textless{}8  & \xmark & C                    & \xmark & \cmark & \cmark & \xmark   \\
MEx \cite{wijekoonMExMultimodalExercises2019}   & 30               & 7          & \textless{}4  & \xmark & S                    & \xmark & \cmark & \xmark & \xmark \\
UP-Fall \cite{martinez-villasenorUPFallDetectionDataset2019}   & 17               & 11         & \textless{}2  & \xmark & F                    & \cmark & \cmark & \xmark & \xmark  \\
Berkeley MHAD \cite{ofliBerkeleyMHADComprehensive2013}  & 12               & 11         & \textless{}2  & \xmark & D,S                  & \cmark & \cmark & \cmark & \xmark   \\
ADL Dataset \cite{dieteVisionAccelerationModalities2019} & 2                & 6          & \textless{}1  & \xmark & D                    & \cmark & \cmark & \xmark & \xmark  \\
ActionSense \cite{delpretoActionSenseMultimodalDataset2022a}  & 9                & 20         & 9$^\ddag$     & \xmark & C                    & \cmark & \cmark & \cmark & \xmark \\
Epic-Kitchens \cite{damenRescalingEgocentricVision2022}   & 37               & 97$^\dag$  & 100           & \cmark & C                    & \xmark & \xmark & \cmark & \xmark \\
Ego4D \cite{graumanEgo4DWorld30002022}  & $\approx$60      & 87$^\dag$  & 160$^*$       & \cmark & D,S,C                & \xmark & \xmark & \cmark & \xmark \\
Ego-Exo4D \cite{graumanEgoExo4DUnderstandingSkilled2024}  & $\leq$550$^{**}$ & 689$^{**}$ & 68$^*$        & \cmark & D,S,C                & \xmark & \xmark & \cmark & \xmark\\
MM-Handwash \cite{burchardImprovedStrategiesMultimodal2026}  & 20               & 2          & \textless{}21 & \cmark & D   & \xmark & \cmark & \xmark & \cmark\\
SAMoSa \cite{mollynSAMoSASensingActivities2022}  & 20               & 26         & 14.2          & \cmark & D,C                  & \xmark & \cmark & \cmark & \xmark \\ \midrule
\textbf{HARMES (ours)}                              & \textbf{20}              & \textbf{15}         & \textbf{61}            & \cmark & \textbf{D,C}                    & \cmark & \cmark & \cmark & \cmark 
\end{tabular}
\end{table}

The table compares the datasets with respect to available sensors, sensor placement, number of participants, duration, and included activities. Some datasets provide additional modalities, e.g., RGB(d)-video data, eye-/gaze-tracking, pose annotations, motion capture point clouds, sEMG, surface pressure, EEG, or passive infrared sensing for presence detection. Some datasets that provide audio and IMU recordings, namely Epic-Kitchens, Ego4D, and Ego-Exo4D, only provide IMU data recorded from a head-worn device, usually recorded alongside their video streams. While these datasets are large, limb movement contains complementary information to egocentric video, whereas head-mounted IMU data decodes mostly the movement of the head \cite{bockWEAROutdoorSports2024}. Datasets that, like HARMES, combine wrist-worn IMUs with audio signals are comparatively limited in size. MHAD \cite{ofliBerkeleyMHADComprehensive2013} contains only 82 minutes of action sequences, and ActionSense \cite{delpretoActionSenseMultimodalDataset2022a} provides 9\,h, while SAMoSa \cite{mollynSAMoSASensingActivities2022} provides a 14.2-hour dataset of 26 activities with single-wrist IMU and audio recordings from 20 right-handed participants. SAMoSa is slightly limited by the placement of a single IMU on the right wrist and by not including any left-handed participants in the experiment group. In comparison, for HARMES, we provide both-wrist IMUs, and we also include three left-handed participants (15\,\%), in line with their estimated share in the world population (9-18\,\%) \cite{papadatoupastouHumanHandednessMetaanalysis2020}. The fully labeled part of HARMES spans more than four times (61\,h) the duration of SaMoSa's, with an additional 20 hours of mostly-labeled diverse free-form activities being provided on top. For activities involving object manipulation and hand interactions, wrist-mounted IMUs provide substantially more discriminative motion signals than head- or torso-mounted sensors. Past research has shown that atmospheric environmental sensors can improve the detection of specific activities, such as hand washing. Specifically including humidity sensors can help detect water- or steam-related activities, and jointly using IMU and humidity sensing can boost classification performance \cite{burchardMultimodalAtmosphericSensing2025,burchardImprovedStrategiesMultimodal2026}. However, the data used in the MM-Handwash study is limited by not including audio data, and by its focus on hand washing, with the only provided labels being \textit{washing hands} and \textit{drying off} \cite{burchardImprovedStrategiesMultimodal2026}. Furthermore, many existing datasets are recorded in controlled or scripted environments, whereas real-world activity recognition requires robustness to variations in execution style and environmental conditions. Our dataset is recorded out-of-lab, in the participants' homes.

However, no multi-class HAR datasets containing atmospheric sensor data exist. Inertial-acoustic datasets are limited in size and/or by their IMU placements. By making available HARMES, our novel dataset with 61 labeled hours and 20 partially labeled hours of both-wrist IMU data, audio data, and data from an atmospheric sensor, we aim to close this research gap and establish our dataset as a novel benchmark dataset for multi-modal activity recognition with these sensors.

\section{Dataset Creation}
\subsection{Study Design}
We recruited 20 adult, able-bodied volunteers for the study. Inclusion criteria were limited to being of legal adult age and having no physical impairments affecting task execution. No specific exclusion criteria were defined beyond the absence of these conditions. In particular, participants were not excluded based on gender, prior experience, or other demographic factors.

All participants were informed about the study protocol and the use of their anonymized data (including publication), and provided written informed consent before participation. The study protocol and data handling procedures were approved by the ethical review board of the University of Siegen (approval no - Ethics: ER\_16\_2025, data handling: 41/2025 VVT), and the study was conducted in accordance with the Declaration of Helsinki.

Table \ref{tab:participants} describes the participant sample. We report the handedness, age, and gender of all participants to enable users of the dataset to analyze or control for effects related to it. We recorded participants from various age groups (range: 18-67) and socio-economic backgrounds, all living in \textit{Germany} at the time of recording the data. We also include three left-handed participants (15\,\%), in line with their estimated share in the world population (9-18\,\%) \cite{papadatoupastouHumanHandednessMetaanalysis2020}. The remaining participants were right-handed. Overall, the participant cohort provides a diverse and representative basis for evaluating activity recognition in everyday settings.

\begin{table}
\centering
\caption{Sample description for the recorded participants. We cover a wide range of ages from the adult population, with participants aged between 18 and 67 (mean: 37.75, sd: 14.4). We recorded a balanced sample of 10 female and 10 male participants. Three out of 20 participants were left-handed.}
\label{tab:participants}
\begin{tabular}{lllllllllllllllllllll}
\toprule
Participant & 01 & 02 & 03 & 04 & 05 & 06 & 07         & 08 & 09 & 10         & 11 & 12 & 13 & 14         & 15 & 16 & 17 & 18 & 19 & 20 \\
\midrule
Age         & 37 & 36 & 34 & 67 & 46 & 47 & 20         & 18 & 51 & 18         & 33 & 62 & 65  & 28         & 33 & 26 & 43 & 32 & 35 & 24 \\
Gender      & m  & f  & f  & f  & f  & m  & m          & f  & f  & m          & m  & m  & m  & m          & m  & m  & f  & f  & f  & f  \\
Handedness  & R  & R  & R  & R  & R  & R  & \textbf{L} & R  & R  & \textbf{L} & R  & R  & R  & \textbf{L} & R  & R  & R  & R  & R  & R\\ \bottomrule
\end{tabular}
\end{table}

\subsection{Recording Setup \& Procedure}
All recordings were conducted by a single researcher who supervised a single participant at a time. Each participant wore a WearOS 5-based smartwatch on their right wrist, a Puck.js\footnote{\url{https://www.espruino.com/Puck.js}} device with an I²C-attached BME280\footnote{\url{https://www.bosch-sensortec.com/en/products/environmental-sensors/humidity-sensors-bme280}} atmospheric environmental sensor on the left wrist. Using sensors on both wrists allows capturing asymmetric and bi-manual interactions, which are common in daily activities and difficult to represent with a single sensor location. The data from the Puck (accelerometer, gyroscope, humidity, temperature, and pressure) was streamed via Bluetooth low energy (BLE) to the smartwatch, which saved it alongside accelerometer, gyroscope, and microphone recordings taken via the smartwatch's internal sensors. In total, we recorded two IMUs (Puck.js, smartwatch), one microphone, and three atmospheric environmental sensor streams (BME280, attached to Puck.js). All firmware artifacts and the source code of our custom-built WearOS application are available on GitHub.
While different hardware platforms are used on each wrist, this setup reflects realistic wearable configurations and introduces variability that can improve model robustness across devices. Adding a second device in a real-world application should be low-cost, and thus a simple device like the Puck.js or a programmable smart band with BLE streaming capabilities should be preferred over asking humans in the real world to wear two smartwatches.

While recording the activities, the researcher who conducted the recording session followed the participant whilst running a Python-based application for instantaneous (on-the-fly) labeling. Instant labeling is preferred, as labeling post-study is both time-consuming and imperfect \cite{kirstenSupervisedLearningDilemma2025}. Participants were asked to conduct a ``clap three times'' synchronization gesture at the beginning of every recording session. This setup made sure that we were able to synchronize all sensor streams, as the clapping is visible in IMU data of both wrists and audible in the microphone recording. As the atmospheric sensors are recorded via the Puck.js, they follow the same clock as the Puck's IMU data and are thus automatically synchronized. The annotator pressed the ``start recording'' button simultaneously with the third clap, so that annotations and sensor streams are synchronous. The combination of real-time labeling and explicit synchronization gestures allows for precise temporal alignment between sensor data and activity labels, minimizing post-hoc annotation errors. The online labels were later cleaned by automated and manual inspection.
For reasons of privacy conservation and anonymity, neither the participant nor the researcher is allowed to talk during the recording sessions. Instead, all activities were carefully explained to the participant before the recording started, and the next activity was silently shown to the participant on the screen of the notebook. The experiment protocol included the possibility of silencing the affected parts of the audio recordings if the participant or any other person could be heard in the recordings. This case occurred a total of seven times, leading to 47.5\,s of audio data being muted by us. This setup allows us to make all recorded data, including the audio recordings, available as part of HARMES without infringing on any person's privacy rights. For details on the muted segments, refer to Section \ref{asec:failures} in the appendix.

During each recording session, participants were tasked with a total of 47 repetitions of 15 different activities (see Section \ref{subsec:activities} for more details). To avoid influencing the participants and ensure a more diverse execution of activities, no exact instructions were given to the participants on how they should conduct each activity. Some activities, e.g., \textit{vacuum cleaning}, have a naturally variable or long duration. For these activities, we did not specify a fixed duration, but to keep them short enough not to dominate the dataset with their duration, we limited the workload to, e.g., one room (cf. Table \ref{tab:activities}).

We chose to record all 20 participants in their homes, leading to a total of 8 different locations, which in turn leads to several benefits in comparison to lab-recorded data. Firstly, the change of environment between participants leads to a high diversity. Different appliances were present, i.e., each tap sounds slightly different, each water-boiler and each vacuum cleaner produces different patterns. Thus, recording in participants' homes introduces substantial variability in environmental conditions and device characteristics (e.g., different taps, appliances, and room layouts), which is critical for training models that generalize beyond controlled laboratory settings. This variability is reflected in the acoustic sensing data, with diverse home appliances producing different sounds, and in the IMU data, with different spatial environments and different appliances being used differently. 
This high diversity prevents the machine learning model from overfitting to specific devices and environments, which would likely have happened if all participants had used the same appliances in a lab. An additional advantage is the familiarity of the participants with their own living arrangements. By recording participants at home, we can rule out certain familiarization effects during the experiments, as participants already know their environment very well, and can conduct each task in a way that feels natural for them. Participants do not need to adapt to previously unknown appliances.

For each recording, our Python-based experiment application generates a randomized permutation of the 47 activity instances in each recording. It then displays a list of these activities, which the experiment then follows. The list contains checkpoints after each set of five activity repetitions, to be able to restart the experiment in case of technical issues or other interruptions.

In between each of the conducted activities, we record short transition periods, which are later assigned to the ``null'' class. However, as this between-activities null class mostly consists of ``standing around and waiting'' and ``walking'' (to the location of the next activity, which might be in another room), we also record one hour per participant of different, free-form activities. The background activities recorded during this hour were chosen not to overlap with any of the activity classes we chose for our dataset. Exceptions to this choice are: \textit{drinking} (in multiple recordings), \textit{washing hands} (Part 16, Rec. 04), and \textit{cleaning table} (Part 13, Rec 04). To avoid erroneous null class classifications in future use cases of the free-form data, we made sure to annotate every occurrence of the selected ADLs in the free-form data. During the one hour, the participants conducted many different complex activities, like baking, building a Lego model, gardening, doing laundry/ironing, petting a dog, drawing, rearranging closets, or solving a puzzle. Participant 13 cleaned their television's screen, which might lead to confusion with window cleaning. For almost all free-form activities, the annotator also recorded start and stop time annotations. These annotations should be viewed as supplementary, as they were not part of the recording protocol and are thus partially incomplete, and the conducted activities differ strongly between participants. 
Most activities in the background class take longer than the short-lived ADLs we focus on for the foreground activities, but in turn, they consist of different complex motion patterns. While we do not explicitly make use of this data during our benchmarking experiments, we verified the integrity and completeness of all recorded data. Potential use cases for this additional hour of data per participant (total: 19.63\,h) could include supplying it as ``null'' examples during supervised training, or using it as a held-out test-set to evaluate false positives, or even using the 20 hours for unsupervised pre-training. One could also use this data for domain adaptation of models that were pre-trained on different datasets. Hence, the mostly annotated background activities provide valuable additional context and can support exploratory analyses, robustness evaluations, and the development of models for more realistic, unconstrained scenarios.

\subsection{Activities}
\label{subsec:activities} 
Table \ref{tab:activities} summarizes the 15 activities each participant was tasked to perform during the main recording sessions. The activities are generally Activities of Daily Living (ADLs), from the categories of personal hygiene and self-care (e.g., washing/disinfecting/creaming hands, brushing teeth), cleaning activities (e.g., wiping the floor, cleaning table, washing dishes, or vacuuming), and preparation/consumption of food/beverages (cutting vegetables, making tea, drinking). By choosing the specific activities listed in Table \ref{tab:activities}, we aimed for a diverse set of real-world ADLs, which are relevant for applications such as monitoring the elderly, assisted living, or general smart home applications. All included activities are prominent ADLs, and we cover multiple subcategories of ADLs. 
Some of the included activities, such as hand-hygiene-related tasks, are additionally relevant for domain-specific applications in areas like food safety and healthcare \cite{mondolHarmonyHandWash2015}.

With HARMES, we aim to provide a benchmark dataset that is explicitly designed to support the study of multi-modal sensor fusion, enabling controlled analysis of how different sensing modalities contribute to activity recognition performance. Specifically, we found that most ADLs can profit from the combination of IMU sensors together with acoustic sensing and environmental sensing as e.g. water-related activities induce humidity changes, kitchen or cleaning tasks produce characteristic sounds. Several selected activities within HARMES are thus difficult to distinguish using inertial data alone (e.g., washing hands vs. washing dishes), but exhibit distinct acoustic or environmental signatures.

\begin{table}
\centering
\caption{A list of all activities recorded for the HARMES dataset. The activities were chosen as highly relevant activities of daily living from multiple categories (self-care, household, feeding). The table also includes a short description of each activity's extent, as well as the number of repetitions of each activity the participants conducted for each of their three labeled recordings.}
\label{tab:activities}
\begin{tabular}{lcl}
\toprule
Activity                & Rep. & Description                                                                                   \\ \toprule
Floor cleaning          & 4    & Wiping: start next to bucket of water, wipe 0.5 - 1 rooms                                             \\ \midrule
Window cleaning         & 4 & \makecell[l]{Applying glass cleaner from spray bottle and wiping \\ using paper towel or cleaning cloth}    \\ \midrule
Vacuum cleaning         &  4 & \makecell[l]{Cleaning the floor with a vacuum cleaner, 1-2 rooms,\\all kinds of vacuums allowed}          \\ \midrule
Washing dishes          & 3 & Washing 3-6 dishes \& cutlery that were prepared in a kitchen basin                                 \\ \midrule
Brushing Teeth          & 3 & Brushing teeth for around 90 seconds, using any kind of toothbrush                            \\ \midrule
Drinking                & 3 & \makecell[l]{Filling and drinking 2-3 sips from a prepared glass of water, putting\\down the glass in between}              \\ \midrule
Apply hand cream        & 3 & Hands were creamed according to the participant's preference                                  \\ \midrule
Putting away dishes     & 3 & Moving dishes from kitchen counter into cupboards, 20-30 pieces                               \\ \midrule
Making tea              & 3 & \makecell[l]{Boil water (volume of 1 cup of tea), addition of \\sugar and/or milk possible but not mandatory} \\ \midrule
Cutting vegetables      & 3 & \makecell[l]{Different kinds of vegetables (e.g. potatoes, apples, carrots),\\cut with a knife on a cutting board.}         \\ \midrule
Disinfecting hands      & 2 & \makecell[l]{Apply disinfectant and rub hands until finished,\\according to participant's preference}         \\ \midrule
Watering plants         & 3 & Fill a watering can, walk to plants and water 2-3 plants in planting pots.                    \\ \midrule
Cleaning table          & 3 & Wipe a table using a wet cloth                                                                \\ \midrule
Cleaning out dishwasher & 2 & Clean out finished dishwasher, usually one of two drawers                                     \\ \midrule
Washing hands      & 4 & \makecell[l]{Wash hands with soap, according to the participant's usual \\hand washing routine, drying off is not included.} \\ \bottomrule
\end{tabular}

\end{table}

\subsection{Data Preparation}

\paragraph{Synchronization and initial Validation}
After recording 20 participants for around four hours each, we manually check each recording for completeness by inspecting the generated recording files and file sizes. After confirmation, the sensor data streams from the different devices must be synchronized. For each sensor, the recordings include a device-time timestamp. As the devices' clocks are not connected, the three-clap synchronization gesture has to be used to align them along the temporal axis. We thus use plots of the accelerometers of the Puck.js, the smartwatch, and the audio data stream, to find ``timestamp zero'', i.e., the timestamp of the third clap, for each recording. This timestamp is then subtracted from all timestamps in the file, and all data with negative timestamps is discarded, as it was recorded before the experiment started. To synchronize the annotations, the experimenter presses the ``start experiment'' button exactly when the participant claps for the third time. This way, all sensor streams are aligned with each other and in sync with the annotations. As another step for initial validation, we visualized and inspected all activity repetitions in the dataset (n=2818) by plotting the sensor data in a vertically stacked time-series plot (provided on GitHub), and could verify a near-perfect alignment for all recordings.

\paragraph{Merging Data of different Time and Sensor Scales}
The in-built IMUs of the Puck.js and the smart watch support slightly different sampling rates, namely 52.5Hz for the Puck.js and 50Hz for the smart watch. Additionally, the Puck.js' accelerometer measures in $\frac{mm}{s^2}$, while WearOS reports the values in $\frac{m}{s^2}$. The BME280 was polled at 1Hz, leading to a significantly lower temporal resolution. As previous work had shown that atmospheric sensors introduce a delay when adapting to changes, a higher frequency would not have added a benefit \cite{burchardMultimodalAtmosphericSensing2025}.

For the IMUs, the sensor data were temporally aligned by resampling to a uniform interval of 20\,ms (50\,Hz) using mean aggregation within each time bin. Missing values introduced during this process were filled via linear interpolation to maintain continuous signals. Following resampling, accelerometer and gyroscope signals were standardized using z-score normalization, resulting in zero mean and unit variance per feature. Normalization was performed independently for each participant and sensor modality, as in the literature \cite{neverovaLearningHumanIdentity2016,mallol-ragoltaHarAGENovelMultimodal2021,zhuangSportRelatedHumanActivity2019}.

The wearable audio data was recorded at 44.1\,kHz, which is a high sampling rate when compared to most other datasets. 
As we took care of possible privacy issues during the design of our experiments, we do not need to apply transformations to anonymize the audio signal and can provide the full raw signal. For further processing with machine learning models, we convert the audio signal into log-mel spectrograms, using librosa \cite{mcfeeLibrosaLibrosa01102025}.

\section{Validation}
\label{sec:validation}
We validate our collected dataset by calculating, reporting, and analyzing several statistics over it. In this section, we also show sample sensor data plots, comparing two of the activities, both in-participant and cross-participant. Additionally, we run initial machine learning and ablation experiments to show that classifiers can be trained with our data and to estimate the impact of each sensor stream on classification performance.

\subsection{Statistics}
Overall, we collected 80.53 hours of recordings from 20 participants (60.90\,h fully labeled) in eight different real-world domestic settings. The mean total recording duration per participant was 241.61 min (fully labeled: 182.71 min). The three labeled recordings per participant amount to roughly three hours, while the partially labeled, free-form background activity recording takes around one hour. The exception is P02, who accidentally hit the ``Stop Recording'' button on the smartwatch after 45 minutes, during the free-form recording.

\begin{figure}
    \centering
    \includegraphics[width=\linewidth]{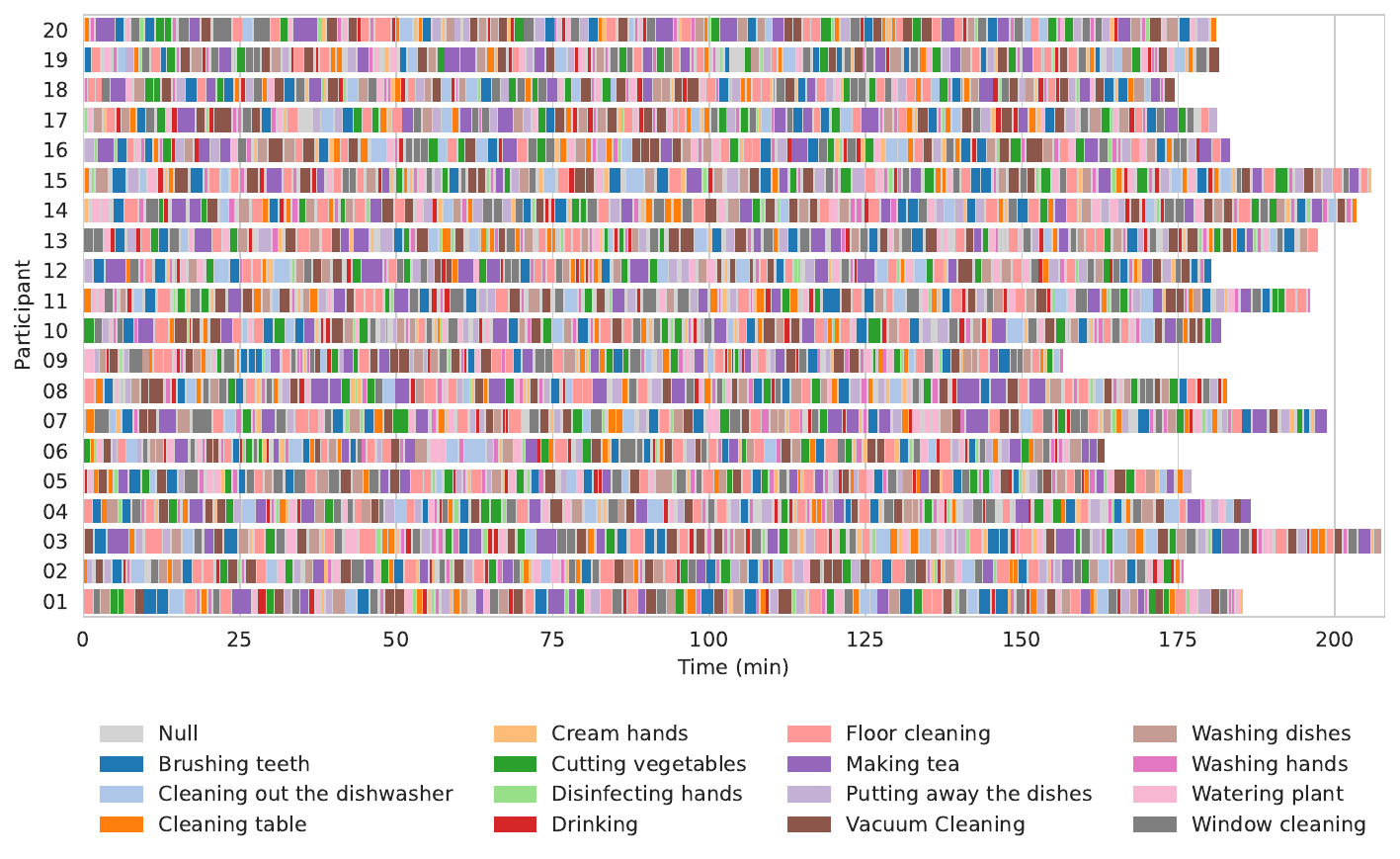}
    \Description{A large overview horizontal barplot with colored sections for each activity. Same activities always have the same color. The plot shows that the order of activities is completely random and that no patterns are visible.}
    \caption{Timeline over all labeled activities, performed by all participants. The three recordings per participant are horizontally stacked. The order of the activities of each recording was randomized before the recording started, to avoid systematic ordering effects such as learning, fatigue, and temporal bias, as well as to reduce correlations between activity type and recording position that could otherwise confound downstream analysis or model training.}
    \label{fig:part_timeline}
\end{figure}

Fig. \ref{fig:part_timeline} displays a per-participant timeline over all three labeled recordings. Each colored section represents a repetition of an activity. To avoid learning effects and other biases, the activity order was randomized separately for each recording and then followed by the participants.

\begin{figure}
    \centering
    \includegraphics[width=1\linewidth]{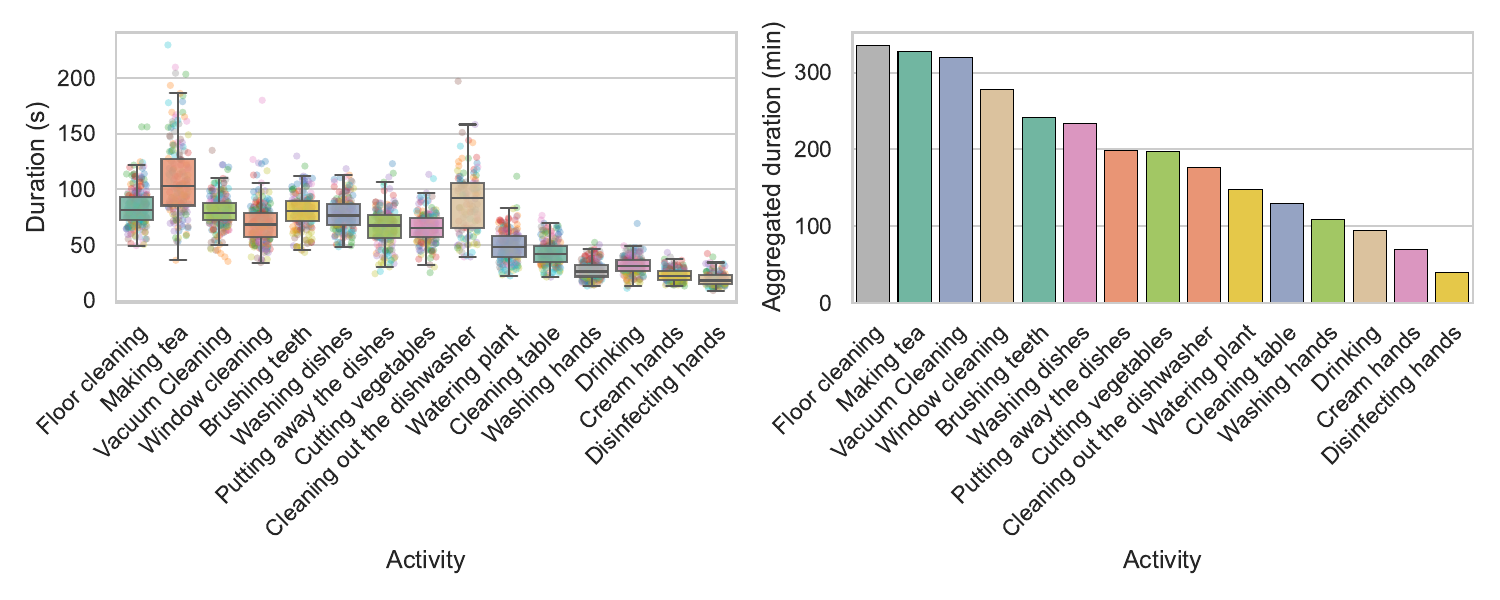}
    \Description{The durations of the activities are distributed relatively similarly. Only making tea has a larger variance and some outliers. On the right side, one can see that the total activity durations are like a staircase, from very long to very short.}
    \caption{Overview over activity durations in our dataset. Left: Boxplots displaying the distribution of the duration of individual instances for each activity, overlayed over scattered dots for each instance, colored per participant. Right: Barplot showing the total duration for each activity in our dataset, aggregated over all participants and instances. The activities are sorted in both subplots from left to right by their total duration in the dataset. }
    \label{fig:act_dur}
\end{figure}

In Figure \ref{fig:act_dur}, we delve deeper by showing the activity durations separately. Figure \ref{fig:act_dur} shows that the durations of each activity can differ between participants and repetitions. Especially \textit{making tea}, which includes waiting for the kettle to finish boiling the water, has a large spread. When it comes to the total durations in the dataset, \textit{floor cleaning}, \textit{making tea}, and \textit{vacuum cleaning} are the most represented, with around 330 minutes each. \textit{Disinfecting hands} is the shortest activity, both by mean instance duration (19.51\,s) and total duration (39\,min). Overall, our dataset contains 15 activity classes, which are all represented reasonably well and by diverse execution styles.

\subsection{Example Data Plots}

\begin{figure}
    \centering
    \includegraphics[width=1\linewidth]{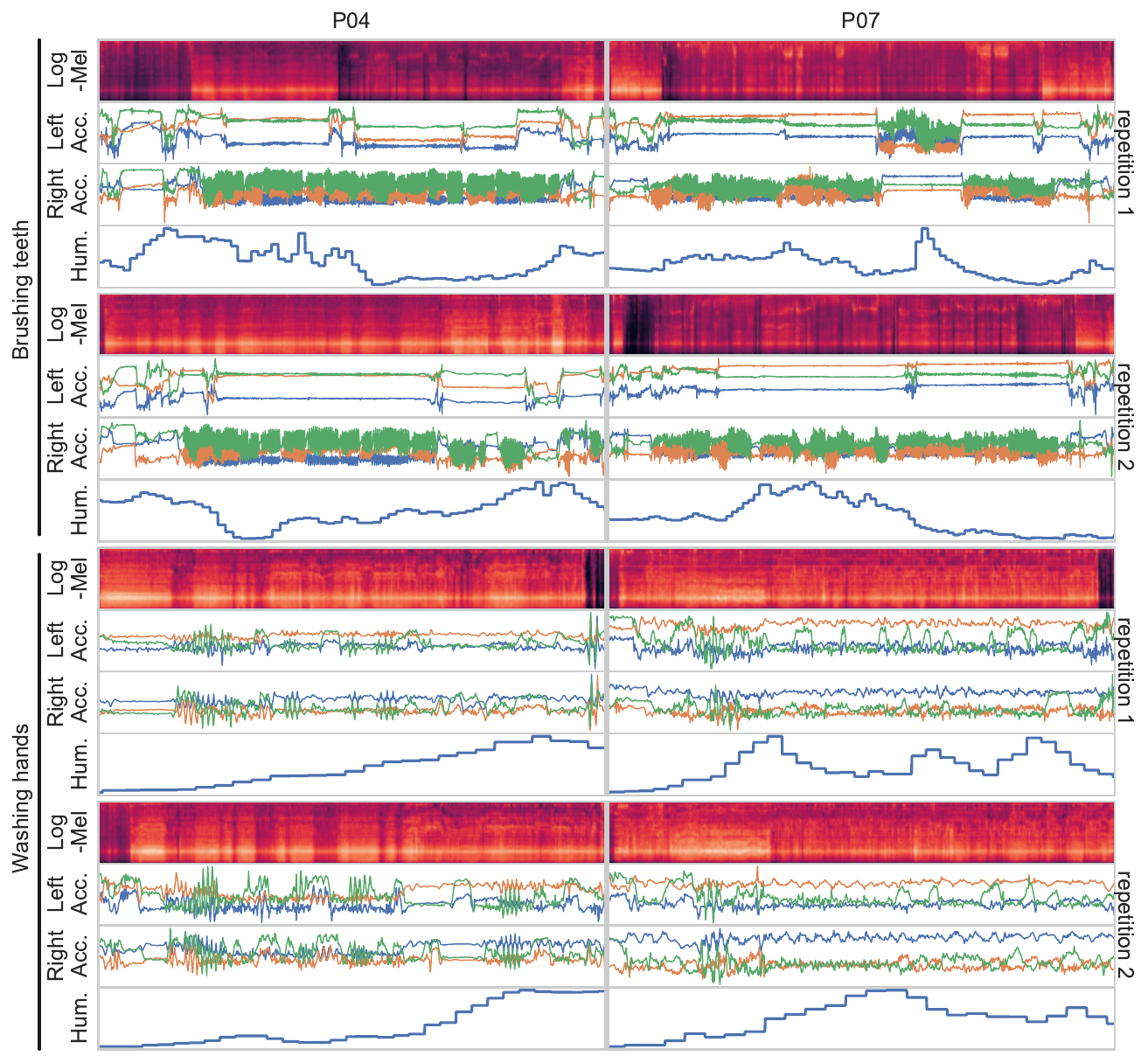}
    \Description{Sample data in the form of audio spectrograms and time-series plots for both IMUs and the humidity sensor.}
    \caption{Exemplary multi-modal data plot, showcasing two instances of \textit{Brushing teeth} (top two rows), and two instances of \textit{Washing hands} (bottom two rows) for both P04 (left column) and P07 (right column). For each instance, we display (from top to bottom) the log-mel spectrogram, the left accelerometer, the right accelerometer, and the humidity sensor. For space-related reasons, we omitted the gyroscopes from the visualization. P07 is one of three left-handed participants included in the dataset. The x-axis relates to the time since the activity started. Note that the x-axes are not equally scaled across the eight subplots, because every subplot shows an entire repetition. For brushing teeth, the durations are all close to 80\,s (P04-1.=80\,s, P04-2.=69\,s, P07-1.=85\,s, P07-2.=91\,s), while for washing hands, the durations are close to 30\,s for all repetitions (P04-1.=27\,s, P04-2.=32\,s, P07-1.=36\,s, P07-2.=25\,s).}  
    \label{fig:large_comp}
\end{figure}

In this subsection, we display some exemplary data plots, highlighting the need for multi-modal approaches to be able to classify the ADLs in our dataset. In Fig. \ref{fig:large_comp}, we show a plot comparing activity repetitions for two activities for two participants, highlighting some noteworthy patterns that are immediately visible in the plots. Firstly, we note that all the different sensor streams are aligned well, showcasing a near-perfect synchronization. Stronger movements are always visible in both IMU signals. If the movement or object interaction creates a loud sound, this is also visible as a vertical band in the log-mel spectrogram. Similarly, the humidity sensor reacts to the proximity to water, e.g., to the tap being turned on. Secondly, the plot shows that for the displayed activities, fairly similar movement patterns can be found intra-participant, while there is a larger difference inter-participant. This aligns well with our observation that participants' activity execution styles differ significantly. For example, participant 07 switched to the left hand for a short duration while brushing their teeth. During hand washing, the tap is turned on, which can be observed well in the log-mel spectrograms.

For further validation, we created similar plots over all activity repetitions for all participants, which we manually inspected to rule out inconsistencies, such as missing data or synchronization errors. The plots display all recorded sensors, with the exception of the temperature and pressure readings. For full transparency, we provide these plots in our GitHub repository as additional materials. Using the plots, we discovered two sensor failures in the dataset, spanning a total of 40 activity instances across two participants (20 each) and described in more detail in Section \ref{asec:failures} in the appendix. We still provide the data, but we excluded it from our downstream machine learning analysis by dropping it when creating the windows.

\begin{figure}
    \centering
    \includegraphics[width=1\linewidth]{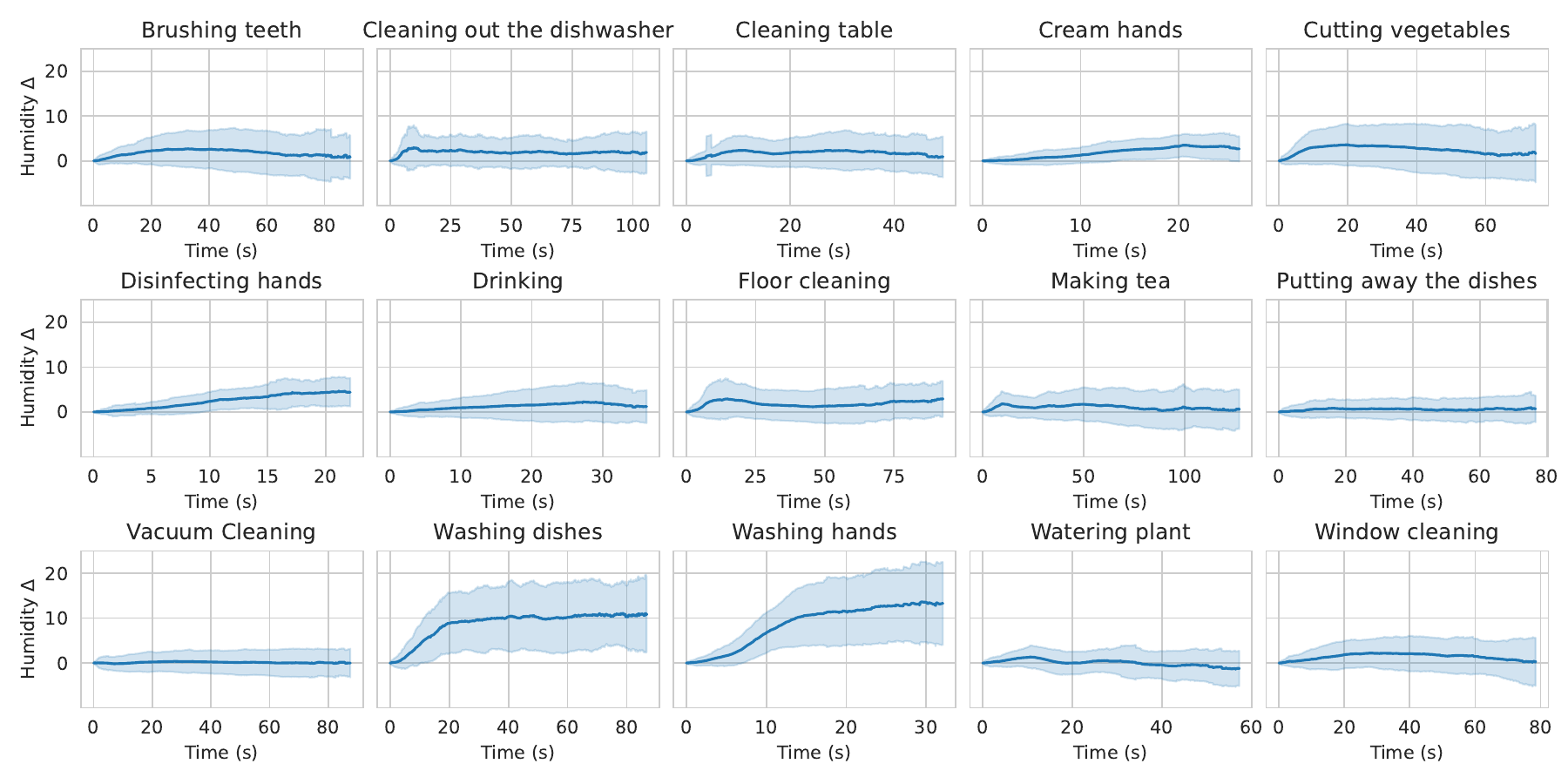}
    \caption{Lineplots per activity class, showing the mean response (solid line) and $\pm 1sd$ (shaded area) of the humidity sensor during the activity. The values are averaged over all instances of all participants in the dataset. The repetitions are cut off after the third quartile of the duration distribution for the respective activity is reached, to ensure enough data is present for meaningful averaging.}
    \label{fig:hum_resp}
\end{figure}

For the inspection of humidity values, we display the average response to the start of an activity in Fig. \ref{fig:hum_resp}. The figure shows the absolute humidity change since the start of the activity (t=0), averaged over all instances in the dataset, with the shaded area displaying $\pm 1\,sd$. For most activities, the response is negligible. For \textit{floor cleaning}, \textit{cleaning out the dishwasher}, and \textit{making tea}, a small bump is visible at the beginning of the activity execution. Disinfecting hands shows a barely noticeable upward trend. For \textit{washing dishes} and \textit{washing hands}, a clear humidity increase of, on average, about $10\,\%$ relative humidity can be observed. 

We inspected analogous plots for the remaining atmospheric sensors (temperature and pressure), but neither exhibited activity-dependent patterns comparable to those observed for humidity under the conditions studied. While barometric pressure has been shown to support the detection of elevation-related activities such as stair climbing or elevator use \cite{karandeRaisingBarOmeterIdentifying2025}, evidence for the utility of ambient temperature in human activity recognition remains limited.

For completeness, both sensors are included in the released dataset, enabling their use in alternative scenarios or future investigations. In this work, however, we focus on modalities that showed clear discriminative characteristics in our exploratory analysis, and therefore do not include temperature and pressure in the subsequent machine learning experiments.

\subsection{Machine Learning and Ablation Experiments}
\paragraph{Experiment Pipeline}
\label{subsec:ml_ablation}
To further validate the collected dataset and to inspect each modality's impact, we run benchmark experiments for both the audio and the IMU data, with established models from their domains. Additionally, we combine all modalities and conduct an ablation study, employing a late-fusion model. We thus propose a baseline deep learning model, inspired by related work and similar to one of the models used by Bhattacharya et al. \cite{bhattacharyaLeveragingSoundWrist2022}. Our model serves as a validation tool for HARMES and as a baseline for future enhancements and more sophisticated architectures. 

\begin{figure}
    \centering
    \includegraphics[width=1\linewidth]{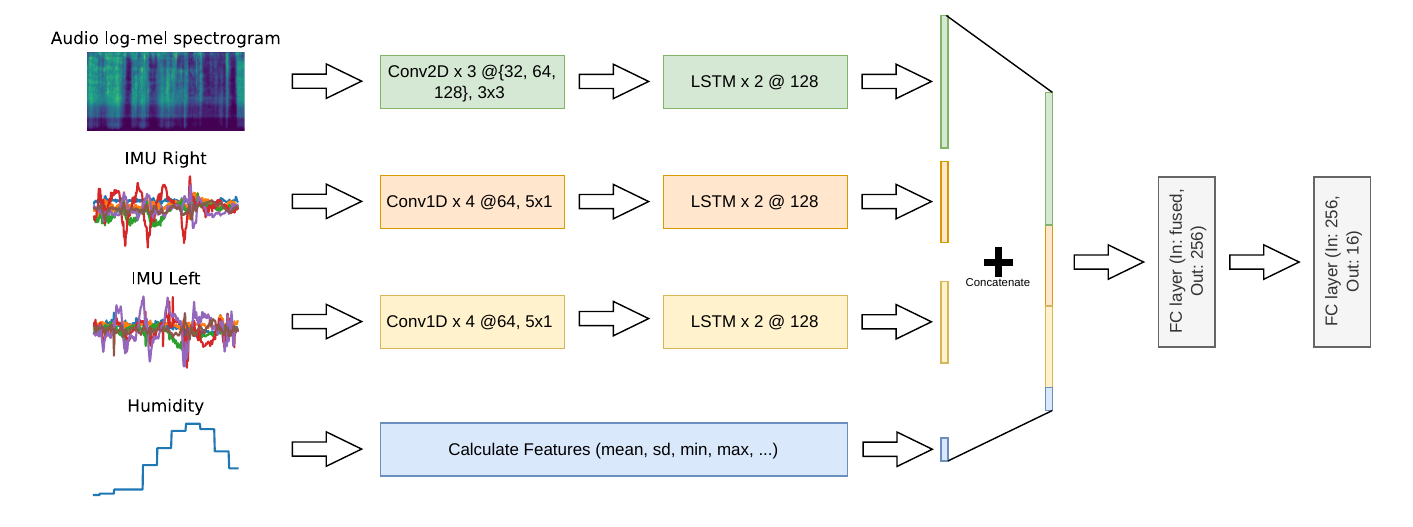}
    \caption{Our multi-modal HAR model. It consists of four branches: The first branch is the audio branch, taking as input log-mel spectrograms, with three convolutional layers followed by two LSTM layers. The next two branches are two identical DeepConvLSTM\cite{ordonezDeepConvolutionalLSTM2016} branches, taking as input the raw IMU (accelerometer + gyroscope) data from the right and left wrist, respectively. The last branch is the humidity branch, which introduces handcrafted features identical to \cite{burchardImprovedStrategiesMultimodal2026}. In the ablation study, branches can be activated or deactivated for the evaluation of their impact. All active branches' outputs are concatenated and fed through two fully connected layers (256, 16) to generate class-wise predictions.}
    \label{fig:model}
\end{figure}

The model architecture is displayed in Fig. \ref{fig:model}. We create two model branches for the audio- and IMU-data. The audio branch consists of three Conv2d encoder blocks and a two-layer LSTM, while the IMU branch consists of two parallel DeepConvLSTM \cite{ordonezDeepConvolutionalLSTM2016,bockImprovingDeepLearning2021} networks (four Conv1d, two-layer LSTM). Similar to Burchard et al. \cite{burchardImprovedStrategiesMultimodal2026}, we use 14 hand-crafted features for the humidity sensor.
Both the audio-branch and the IMU-branch output a 256-dimensional embedding (128 dimensions per wrist IMU), while we arrive at a total of 14 dimensions for the humidity features. These embedding features are concatenated and fed through a two-layer (256,16) fully-connected classification head.

We train the models both on 5\,s and 10\,s long, non-overlapping majority-vote-labeled windows, using the ADAM optimizer and weighted cross-entropy as the loss function, for 35 epochs with a batch size of 32. Implementation details of the chosen training pipeline can be found in Section \ref{asec:ml_params} in the appendix.
All experiments are conducted in a leave-one-participant-out (LOPO) fashion to best approximate the model performance on a completely unseen participant.

First, we train and evaluate models on each modality individually to study each sensor's strengths and ambiguities separately. We then evaluate almost all potential combinations of sensors to estimate their influence and importance for the classification results. This ablation study includes LOPO runs for which we left out the audio, one or both IMU signals, and/or the humidity signal. 

\paragraph{Results}
Table \ref{tab:results} shows all the tested combinations and their participant-averaged classification performances, for both 5\,s and 10\,s windows. For per-participant results, we refer the interested reader to Table \ref{atab:all_results} in the appendix.

\begin{table}
\caption{Multi-modal model performance by sensor configuration and window size. The sensor configurations cover relevant sensor combinations (IMU\_L/R~=~IMU on \textbf{L}eft or \textbf{R}ight wrist, A~=~Audio, H~=~Humidity, ALL~=~All sensors). We report accuracy and F1 score (macro and class-weighted), for both 5\,s and 10\,s windows. The chance level of a stratified dummy classifier reaches a macro F1 of 0.06 for both window sizes.} 
\label{tab:results}
\begin{tabular}{lrrrrrr}
\toprule

Window Size & \multicolumn{3}{c}{5\,s} & \multicolumn{3}{c}{10\,s} \\ \cmidrule(lr){2-4} \cmidrule(lr){5-7} 
Sensor Config  & Accuracy & F1 (macro) & F1 (weighted) & Accuracy & F1 (macro) & F1 (weighted) \\  
\midrule
H & 0.143 & 0.098 & 0.117 & 0.150 & 0.112 & 0.121 \\
IMU\_L & 0.475 & 0.456 & 0.473 & 0.494 & 0.473 & 0.490 \\
IMU\_R & 0.599 & 0.570 & 0.598 & 0.623 & 0.598 & 0.622 \\
IMU\_L+IMU\_R & 0.639 & 0.619 & 0.639 & 0.652 & 0.628 & 0.651 \\
IMU\_L+IMU\_R+H & 0.638 & 0.617 & 0.638 & 0.634 & 0.609 & 0.634 \\
A & 0.738 & 0.700 & 0.738 & 0.749 & 0.708 & 0.749 \\
A+H & 0.736 & 0.696 & 0.735 & 0.756 & 0.717 & 0.757 \\
IMU\_R+A & 0.771 & 0.736 & 0.772 & 0.781 & 0.750 & 0.783 \\
IMU\_L+IMU\_R+A & \bfseries 0.794 & \bfseries 0.763 & \bfseries 0.795 & \bfseries 0.791 & \bfseries 0.760 & \bfseries 0.794 \\
ALL & 0.789 & 0.754 & 0.789 & 0.789 & 0.758 & 0.791 \\
\bottomrule
\end{tabular}
\end{table}

\begin{figure}
    \centering
    \includegraphics[width=1\linewidth]{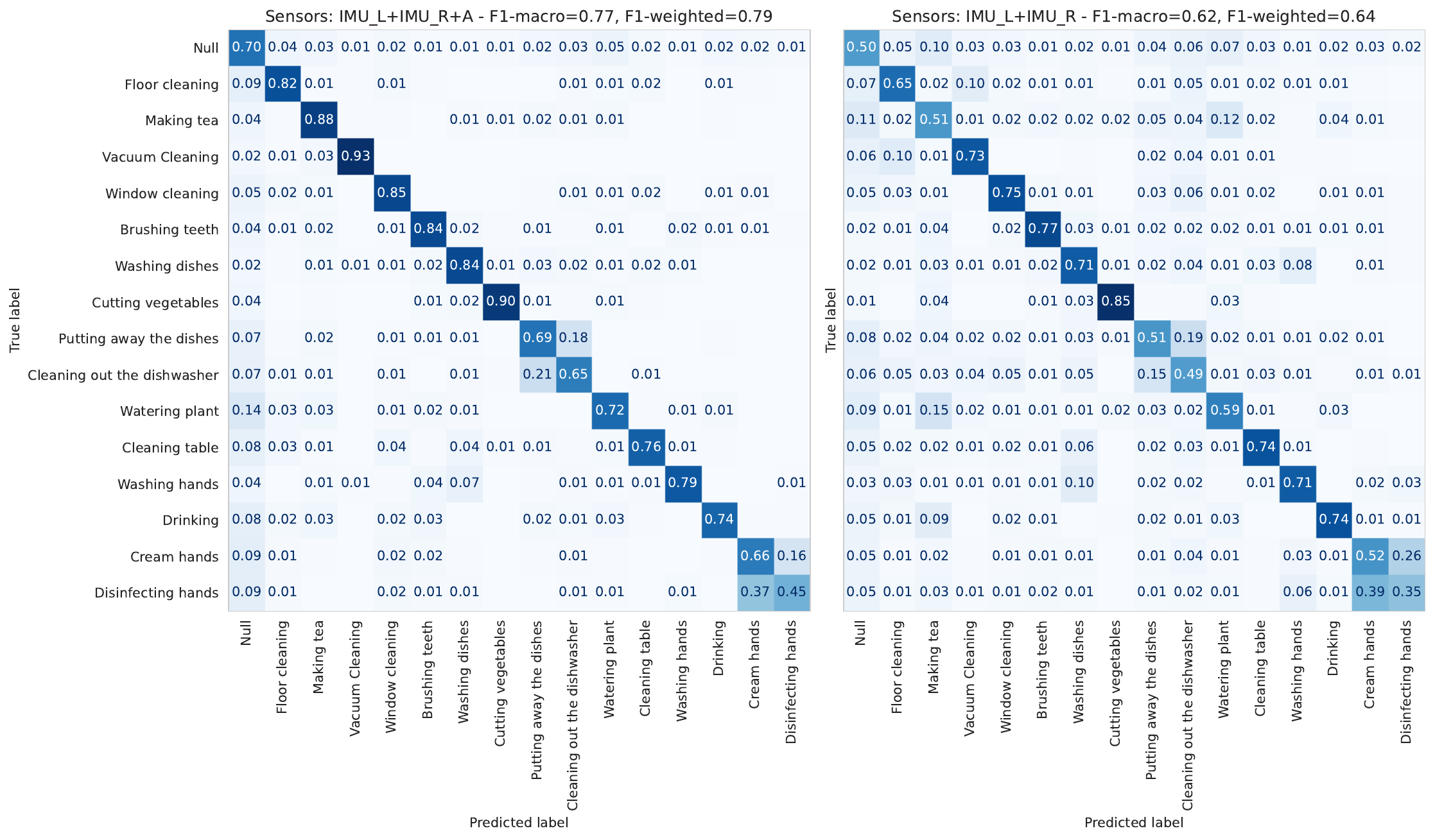}
    \caption{Confusion matrices for the activity classification tasks (5\,s windows),  for the combination of IMU sensors and audio data (left) and for IMU sensors only (right). The confusion matrices are row-normalized (i.e., over the true labels). Values below 0.005 are omitted for better readability. Note that while the performance is improved for the multi-modal model for all classes, especially the Null-class, and often confounded activities like \textit{floor cleaning} and \textit{vacuum cleaning} or \textit{putting away the dishes} and \textit{cleaning out the dishwasher} profit significantly, highlighting the complementarity of audio and inertial sensing.}
    \label{fig:conf}
\end{figure}

The results are similar for both 5\,s and 10\,s windows. We report accuracy, macro F1-score, and weighted F1-score as the main metrics. The performance difference between the two window sizes is minimal. We report and discuss the 5\,s results, as they are marginally better. The weighted F1-score is generally slightly higher than the macro F1-score, hinting at more frequent misclassifications for the less represented classes, as we can confirm through the analysis of confusion matrices (cf. Fig. \ref{fig:conf}). The macro F1-score assumes equal importance for each class and is thus our metric of choice for the analysis of the models' classification performance. Humidity as a singular modality does not suffice to accurately predict the current activity. While the model trained only on humidity data outperforms the stratified random baseline classifier (F1:$0.06$), it only reaches an F1 score of $0.1$. However, the confusion matrices (cf. Fig. \ref{afig:conf_h} in the appendix) show that especially \textit{washing dishes} and \textit{washing hands} can be detected with accuracies of 30\,\% to 40\,\%, while often being confused with each other. In line with our findings from the visual inspection in Fig. \ref{fig:hum_resp}, this confirms that the humidity signal provides useful contextual information for the detection of these classes.
The results of the IMU-based classification models are consistent with those from previous work. As 17 of 20 participants in the sample are right-handed, the right-wrist IMU provides a more informative signal to the model than the left-wrist IMU. The combination of the two IMUs yields slightly better classification performance. However, the reached F1-scores of around $0.62$ are relatively low, especially when compared to the macro F1-score of $0.70$ reached by the audio-only model. The combination of IMU from the right or from both wrists with the audio data yields another performance increase to $0.74$ and $0.76$, respectively. These results underline the complementarity of inertial and acoustic sensing for activity recognition. However, the results remain almost unchanged when the models additionally have access to the humidity sensor features, across both window sizes. While we showed a visible response in the humidity signal for water-related activities in Fig. \ref{fig:hum_resp}, this did not translate into a measurable performance increase. Rather, the performance slightly decreases when humidity features are available, to $0.75$. Possible reasons for this gap could be the previously described delay and also the windows being too short for the specific sensor type. As the response in Fig. \ref{fig:hum_resp} is strong, we still expect that the HARMES dataset can be employed to develop and validate better integrations of humidity as a modality into multi-modal HAR models.

Fig. \ref{fig:conf} shows the confusion matrix for the best-performing sensor and window size combination (left, both IMUs and audio, 5\,s windows), as well as for the IMU-only model (right). The confusion matrices for all tested sensor- and window size combinations can be found in Section \ref{assec:confusion} in the appendix. Notably, semantically similar classes like \textit{putting away the dishes} and \textit{cleaning out the dishwasher}, as well as \textit{cream hands} and \textit{disinfecting hands}, are often confused with each other. These classes are challenging because their motion and sound patterns overlap significantly. If researchers wanted to remove this ambiguity from the dataset, they could simply combine these two pairs of labels into one new label each. These activities might also be seen as an opportunity for context-based HAR models, more so than ``only'' models profiting from the availability of multi-modal data. These four classes are also the worst-performing in the dataset, with \textit{cream hands} and \textit{disinfecting hands} being the two smallest classes, which makes it even more difficult for the model to correctly predict them. Overall, the comparison of the confusion matrices shows an improvement across all classes for the audio-augmented model. Especially, \textit{making tea}, which is a low-motion class and was often confused with \textit{watering plant}, can be predicted more accurately. Audio data also helps reduce the ambiguity between \textit{cream hands} and \textit{disinfecting hands}. Similar ambiguities between \textit{floor cleaning} (wiping) and \textit{vacuum cleaning}, and between \textit{washing hands} and \textit{washing dishes} are also mostly eliminated by the inclusion of audio data.

\begin{figure}
    \centering
    \includegraphics[width=1\linewidth]{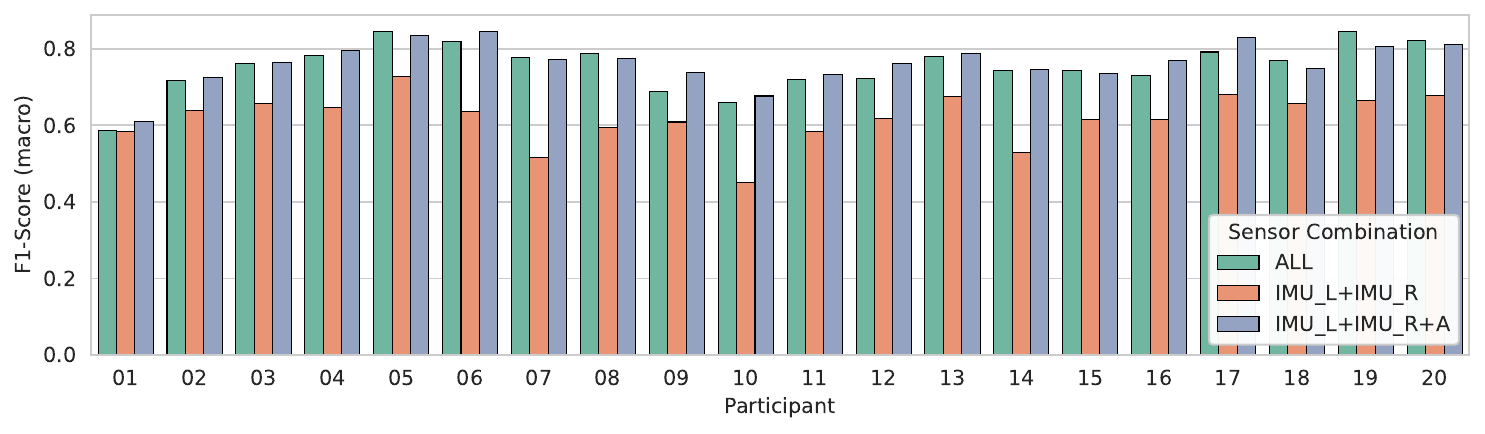}
    \caption{Barplot showing per-participant macro F1-score results of the LOPO cross-validation (5\,s windows), for IMU sensor only models, for the best sensor configuration (IMUs and audio), and for all sensors. While IMU-based models struggle with the left-handed participants (07, 10, 14), the addition of the audio modality leads to improved performance for every left-out participant.} 
    \label{fig:res_per_part}
\end{figure}

In Fig. \ref{fig:res_per_part}, we display the macro F1-score results of the LOPO cross-validation on 5\,s windows for IMU sensor only models, for the best sensor configuration (IMUs and audio), and for all sensors. The plot shows that the addition of the audio modality and humidity sensor leads to a performance increase for every participant. We also observe a reduced variance (sd=0.05 vs. sd=0.08 for IMU-only). While the IMU-only model struggles with the left-handed participants (07, 10, 14), the performance on these participants is aligned with the other participants when audio data is included.

The window sizes of five and ten seconds are similar to the literature \cite{bhattacharyaLeveragingSoundWrist2022}, but smaller windows are also often viable \cite{bockWEAROutdoorSports2024}, at least for IMU-based HAR.
We avoided shorter windows because the humidity sensor only measures a relative effect. Its rate of change over time is much more relevant than its absolute value. In a shorter window, this could not be measured at 1Hz. On the other hand, future work can explore whether shorter windows are also a feasible option for audio-inertial HAR models on our dataset.

Overall, the machine learning performance evaluation shows that employing multiple modalities from our HARMES dataset leads to better models that generalize well to unseen participants. While models based on singular modalities reach F1 scores between $0.46$ and $0.7$, the combination of multiple sensor streams yields significant advantages for the classification performance. While the multi-modal model performance varies between subjects, it remains above a macro F1 score of $0.61$ for all of them, reaching a maximum of $0.84$, while averaging $0.76$ (median $0.77$). This validation underlines that our dataset is synchronized well and that complementary information is contained in the different sensor streams. The validation also corroborates prior findings that multi-modal data is crucial for enabling state-of-the-art HAR performance.

\section{Discussion}
As for all datasets in HAR, collecting data is a time-consuming task that requires substantial effort. Collecting enough data is, however, crucial for training deep-learning models and for enabling proper generalization. With the results for the classifiers trained on our dataset in a leave-one-participant-out paradigm, we can show that our dataset is indeed large enough to learn most activities well.
From the results of the ablation study, we conclude that combining IMU sensors and acoustic sensing leads to a consistent and measurable improvement in performance ($0.76$ for IMUs with audio vs $0.7$ for audio-only and $0.62$ for IMUs-only), with the combination of both-wrist IMUs and audio data achieving the highest macro F1 score of $0.76$. Activities that are difficult to distinguish from a single modality alone, like \textit{watering plant} or \textit{making tea}, profit the most. Our dataset would also lend itself well to studies that are aimed at comparing different sensor combinations more in-depth and to studies evaluating sensor-fusion strategies. 

However, the addition of the humidity sensor did not have the desired positive impact on the classification performance when we applied methods from the literature. Possible reasons for this are manifold. First of all, the features we calculated are of lower dimensionality than the IMU and audio branches (14-dim vs. 256-dim each). Second, even five- and ten-second-long windows do not perfectly capture the longer-lasting temporal trends we observed when visualizing the humidity data. E.g., for hand washing, the humidity peak is only reached after around 20\,s, possibly implying that humidity sensing can only be used to provide context in a larger time frame. Moreover, humidity as a standalone modality performs poorly when the model is trained exclusively on it. However, it shows comparatively better performance for water-related activities such as \textit{washing hands} and \textit{washing dishes}. The HARMES dataset is the first to make available such atmospheric sensor data along with IMU and audio data, enabling the community to develop models that can better extract the contextual information contained in the atmospheric sensor signals, while at the same time making use of IMU and audio data.

The activity classes for which the models trained on the dataset struggle the most are \textit{Null} and \textit{cleaning out the dishwasher}/\textit{putting away dishes}, as well as \textit{cream hands}/\textit{disinfecting hands}. For the latter two combinations, the confusion matrix clearly shows that the model is getting confused by the similarity between the two classes of each pair. \textit{Cleaning out the dishwasher} has a large overlap with \textit{putting away dishes}, as both activities consist of the participant moving the dishes from the starting location (dishwasher/kitchen counter) to their location in a shelf. Similarly, the movements of \textit{cream hands} and \textit{disinfecting hands} are strongly related, with the only larger difference being the product used on the hands. These confusions highlight a fundamental limitation of sensor-based HAR, namely that activities with highly similar motion patterns and object interactions remain difficult to distinguish, even with multi-modal sensing.

\subsection{Limitations}
One limitation of the HARMES dataset is that it contains no speech at all. In a more realistic scenario, humans could be heard talking, possibly masking activity-related sounds and thus making it harder to spot activities. When using a dataset with no speech, robustness against this type of environmental noise is likely not learned as much by the classifiers. However, it would be relatively simple to artificially add speech sounds or other noises to the audio tracks, so that this limitation could be circumvented. The audio modality could also face issues in the real world, when other loud noises overshadow the activity-specific sounds, e.g., someone else could use a vacuum nearby, and this might confuse models.

Another limitation is the use of two separate and different wearable devices, which may not reflect typical consumer setups and could introduce additional variability across sensors. Furthermore, while the dataset includes 20 participants, larger cohorts would be beneficial to further improve generalization. Additionally, the dataset focuses on indoor, home-recorded ADLs and thus may not generalize to outdoor or industrial settings. Finally, the semi-naturalistic data collection does not allow us to measure real-world in-the-wild performance. For doing so, a completely unscripted dataset would have to be collected and labeled, requiring significant additional effort.

\subsection{Future Work \& Dataset Use Cases}
By applying open licenses to the data and publishing all code, we hope to encourage other researchers to use our dataset for all purposes imaginable. We supply 61 hours of fully labeled and 20 hours of additional multi-modal, mostly labeled data from other activities. The significantly larger size of the dataset, paired with its 20 participants and diverse locations, can be utilized to train models that generalize even better to unseen users and environments. The inclusion of the mostly labeled background data as the ``Null'' class can also help explicitly tackle the ``Null class problem'' \cite{bullingTutorialHumanActivity2014}, which is a necessary trait for in-the-wild systems.
Due to HARMES being significantly larger in size than related datasets, application of more sophisticated deep learning models, such as ones that include self-attention layers or transformers, as well as self-supervised learning approaches (SSL), could be investigated. We hope to establish the HARMES dataset as one of the communities' go-to benchmark datasets for multi-modal HAR models. As a well-synchronized multi-modal dataset, it lends itself especially well to studies on sensor-fusion (e.g., comparisons of early-, late-, and gated fusion).

HARMES can also be employed to answer the open question of whether adding a second wrist-IMU provides a good return on investment. While the performance was slightly higher with the second IMU in our baseline experiment, single-IMU models together with audio recordings performed almost as well, hinting that a single smartwatch containing both sensor types could be sufficient for the highly reliable classification of the ADLs in our dataset. This finding suggests that combining audio with a single wrist-worn device may provide a favorable trade-off between sensing complexity and performance, which is particularly relevant for real-world deployments. Future work can further evaluate this finding. 

Future work could also investigate adaptive sensing strategies, where modalities are selectively activated depending on the activity context, enabling more energy-efficient wearable systems. A student-teacher approach similar to Liang et al. \cite{liangAudioIMUEnhancingInertial2022} could also yield an energy-efficient model while profiting from the initial availability of synchronized audio data for training.

We also plan to further explore how the atmospheric sensor signals can be better incorporated into the activity classification pipeline, as their impact in the ablation experiments was negligible, though visual inspections showed clear patterns. One possible avenue would be the addition of humidity as context, in a more sophisticated way. Shorter activity windows could be augmented with longer-term humidity trends, i.e., a 10\,s window of IMU data could be augmented with access to the last 100\,s of humidity data, or trends derived therefrom. Overall, our evaluation showed that the state-of-the-art integration of atmospheric sensing in HAR is still in need of future research, as the clearly measurable signal response is not reflected in improved HAR performance yet.

\section{Conclusion}
We present HARMES, a 20-participant, 80-hour, multi-modal labeled dataset for HAR with wrist-worn data from two IMUs, high-resolution audio recordings, and atmospheric sensors. By including synchronized dual-wrist IMUs, wrist-worn microphone recordings, and data collected from atmospheric sensors, our dataset provides a combination of modalities not previously present in a public dataset, while maintaining high ecological validity by recording a diverse sample of 20 participants in their homes. HARMES contains almost six times as many hours as the previously largest state-of-the-art wrist-inertial acoustic dataset (SaMoSa), while contributing data from an additional wrist IMU. It therefore fills the gap of a large-scale labeled inertial-acoustic dataset for HAR. 

Furthermore, we describe and validate the collected dataset via visual inspection, by calculating statistics, and by training baseline machine learning models. We show that models which generalize to unseen participants can be trained with both IMU and audio data separately, and that the combination of sensing modalities yields an increase in classification performance. We run an ablation study to further inspect each sensor's contribution. We find that the use of audio and IMU sensors in a joint model boosts classification performance significantly. From visual inspection and our ablation study, we realize that while humidity shows a distinct reaction in the signal for water-related activities, it is likely more useful as a slow contextual latent variable, providing an opportunity for further exploration. Our findings confirm that acoustic sensing provides complementary information to inertial measurements, particularly for activities that are ambiguous from motion data alone. 

The dataset, code, and data collection applications are made available in a fully open-source fashion, enabling other researchers to easily access and use the data without restrictions. Overall, HARMES enables the systematic study of inertial-acoustic multi-modal sensing for human activity recognition and demonstrates that combining motion, acoustic, and environmental signals improves robustness for challenging real-world activities.

\section*{Author Contributions}
R.B. co-designed the study, prepared all recording hardware, firmware, and wearable recording apps, co-developed the recording procedure, oversaw the recording process, ran all validation experiments, and wrote the manuscript.
P.B. co-designed the study, wrote the annotation and synchronization tool, recruited all participants, conducted the recordings, and synchronized the recordings.
M.B. co-designed the machine learning validation experiments and co-wrote parts of the manuscript.
All authors reviewed the manuscript.

\begin{acks}
The authors would like to thank all participants.
We gratefully acknowledge the University of Siegen's OMNI cluster.  
\end{acks}

\bibliographystyle{ACM-Reference-Format}
\bibliography{bibliography_better_bibtex}

\appendix

\section{META - Methodological Transparency}
\label{asec:meta}
\subsection{Dataset and Code Availability}
\label{assec:data_code_avail}
We make available raw and unprocessed recordings, along with cleaned, processed, and prepared data at Zenodo \cite{burchardHARMESMultiModalDataset2026}. We describe the data format in Section \ref{assec:data_format}. The dataset is archived on Zenodo with a persistent identifier to ensure long-term availability. We publish all code in a GitHub repository \textit{\url{https://github.com/RBurchard/HARMES}}. The code includes all recording tools (i.e., Puck.js firmware, Wear OS recording application, live-annotation tool), as well as all code used for merging, resampling, normalizing, and synchronizing the data, and all code used in the validation (i.e., data loading, processing, and machine learning experiments). Both the code and the dataset repository contain extensive ReadMe files, in which we explain their composition and instruct the reader on how to use the data and run the experiments.

We license all code and data under open licenses, applying the MIT license to the code and CC-BY to the data.

The availability of all collected data and the code for our entire pipeline supports not only the full reproduction of our downstream experiments, but also allows for an independent replication of our data collection by other researchers with novel participants. The provided codebase enables end-to-end reproduction of our experiments, including data loading, pre-processing, and model training. This lowers the barrier to entry and facilitates direct comparison with future work.

\subsection{Data Format}
\label{assec:data_format}
We supply the raw recordings in the following structure: For each participant, there exists a folder, and the participant's folder contains the four recordings for the participant. Recordings one, two, and three are labeled, while recording four is free-form and mostly labeled. Recordings that are split into two parts contain both parts as separate subfolders. In turn, each recording's folder contains separate files for the data from the Puck.js (IMU, BME280, .csv), the watch IMU (.csv), and the audio data (.h5, HDF5-format, key~=~"Audio"). Additionally, a file containing all labels (.csv) is present in each folder. All .csv files contain a column called \textit{ts\_sync} which contains a synchronized timestamp in milliseconds, since the third peak of the synchronization gesture. The audio data has a 44.1 kHz sampling rate, i.e., 44,100 samples per second, and starts at $ts_{sync} = 0$.

Additionally, we supply a ready-to-use pre-processed version containing all data. We supply all participants as separate Python pickle files. The data in these files is already synced, resampled (50 Hz for IMU data), and z-score normalized. By loading the files from Python with a single command, researchers can immediately use the data to run downstream experiments, without the need for any custom data loading code or additional pre-processing.
The structure of the pre-processed version is as follows:
\begin{samepage}
\begin{itemize}
    \item Dataset (stored as one Python pickle file per participant):
    \begin{itemize}
        \item \textbf{Filename}: \textit{participant\_<Participant ID>.pkl}
        \item \textbf{Content}: List containing the participant's recordings as dictionaries.
        \begin{itemize}
            \item Each participant has \textbf{four/five recordings}
            \item Each recording is a dictionary with the following keys:
            \begin{itemize}
                \item \texttt{IMU\_L} (left IMU data, 50\,Hz)
                \item \texttt{IMU\_R} (right IMU data, 50\,Hz)
                \item \texttt{Audio} (audio signal, 44.1\,kHz)
                \item \texttt{BME280} (environmental sensor data, 50\,Hz)
                \item \texttt{Labels} (Annotations with start/stop times in ms)
                \item \texttt{rec\_id} (ID of the recording, same as the folder name of the recording in the raw version.)                
            \end{itemize}
        \end{itemize}
    \end{itemize}
\end{itemize}
\end{samepage}

\subsection{Machine Learning Details and Parameters}
\label{asec:ml_params}
We report all relevant training details and chosen hyperparameters in this section. All models were implemented, trained, and tested in Python $3.11$, using the PyTorch\footnote{\url{https://pytorch.org}} framework. We use window sizes of 5\,s and 10\,s, and a Leave-One-Participant-Out cross-validation. The models are trained for 35 epochs. We selected a class-weighted cross-entropy loss as the loss function to address the class imbalance of the dataset. The class weights $\textbf{w}$ are calculated from the training data of each split by first counting each class $C_i$'s support in the training data ($N_i$, see Eq.\ref{eq:ni}) and then calculating the weight $w_i$ for each class as the inverse of the support, with the addition of a small $\epsilon$ value to rule out division by zero (Eq. \ref{eq:wi}). The weights are then normalized (Eq. \ref{eq:w}) and multiplied by the total number of classes (16).
\begin{align}
N_i &= |\{x | x \in C_i\}|\label{eq:ni}\\
w_i &:= \frac{1}{N_i+ \epsilon}, \epsilon = 10^{-6} \label{eq:wi}\\
\textbf{w} &:= \frac{\textbf{w}}{\sum{w_i}} \cdot |C| \label{eq:w}
\end{align}
These weights are then passed to the loss function, which adapts the loss for each class according to the provided weights.
We train the models with mini-batches of size 32, using ADAM as the optimizer with a fixed learning rate of $0.001$. We use dropout only for the fully connected layers in the classification head, with $p=0.3$.
For implementation details, we refer the interested reader to our code repository, linked in Section \ref{assec:data_code_avail}.

The 14 hand-crafted statistical features used for including humidity are: mean, sd, min, max, median, range, inter-quartile-range, mean diff., sd diff, max. absolute diff., slope, mean crossings, fraction of values above mean, energy. 

\section{Additional Dataset Description and Notes}
\subsection{Recording errors, muted sections, and missing data}
\label{asec:failures}
For participant 05 recording 01, part 01, most of the humidity data is missing (for exactly 20 activity instances), and for participant 14, recordings 01, part 01, and recording 03, the BME sensor seems to have had an unstable connection to the Puck.js, so that for a total of 20 activity instances, the signal is at least partially missing.

Recordings could be split into parts if a technical or human error occurred. We made use of this feature for nine out of 80 recordings, which had to be split into two parts each. When the experiment is paused and restarted, the synchronization gesture is re-performed by the participant, and the next activity after the last fully reached checkpoint is selected as the current activity.
Reasons for restarting were: Disconnection issues between Puck.js and Smartwatch, unstable connection to the BME280 sensor attached to the Puck.js, and errors with the activity order.

In addition to the data we excluded with missing sensor values, for two participants, a single activity repetition was simply forgotten to record, by accidentally skipping over it during recording. Namely, one instance of \textit{washing hands} is missing for participant 11, and one instance of \textit{putting away the dishes} is missing for participant 16. All other recordings are complete w.r.t the activity repetitions.

The study protocol requires us to mute segments of the audio file if voices can be heard. We had to do so for a total of seven times, with the muted duration summing up to 47.5\,s. Luckily, only 1.5\,s of these were during activity executions of the labeled recordings, namely 0.5\,s of recording 0702 (during \textit{brushing teeth}) and 1\,s of \textit{window cleaning} in 1902, part 01. Additionally, we muted parts in between activity executions or in the free-form recordings for the recordings 0502 (14\,s), 1703 (25\,s), 2004 (7\,s)

\subsection{Additional statistics}
\begin{figure}[H]
    \centering
    \includegraphics[width=0.5\linewidth]{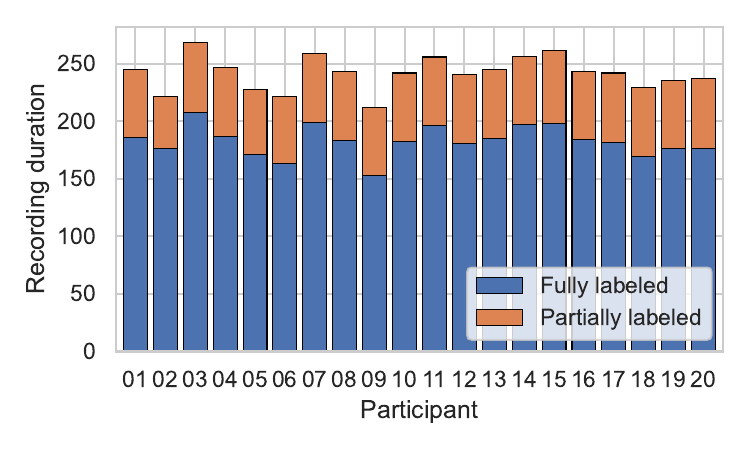}
    \caption{Recording duration per participant, split into the combined duration of the three supervised sessions (labeled) and the $\sim$1\,h free-form activities recording, where most activities are labeled too. Total recording durations depend on the task execution speeds of the participants. While we aimed for one hour per recording, slightly longer or shorter recordings were possible.}
    \label{fig:part_dur}
\end{figure}

Fig. \ref{fig:part_dur} shows the amount of data collected for each participant, split into the fully labeled, structured recordings and into the free-form, mostly labeled recording time. The free-form activities are mostly labeled in the form of comments in their respective label files, but they do not contain the activities we chose for the structured recordings of the dataset.

\newpage

\section{Machine Learning Results}
\subsection{Confusion Matrices}
\label{assec:confusion}
This section lists the confusion matrices of every model we tested, for each combination of applied sensors, as well as for both window sizes (05\,s and 10\,s, IMU\_L/R~=~IMU on \textbf{L}eft or \textbf{R}ight wrist, A~=~Audio, H~=~Humidity, ALL~=~All sensors).

\begin{figure}[H]
\includegraphics[width=\linewidth]{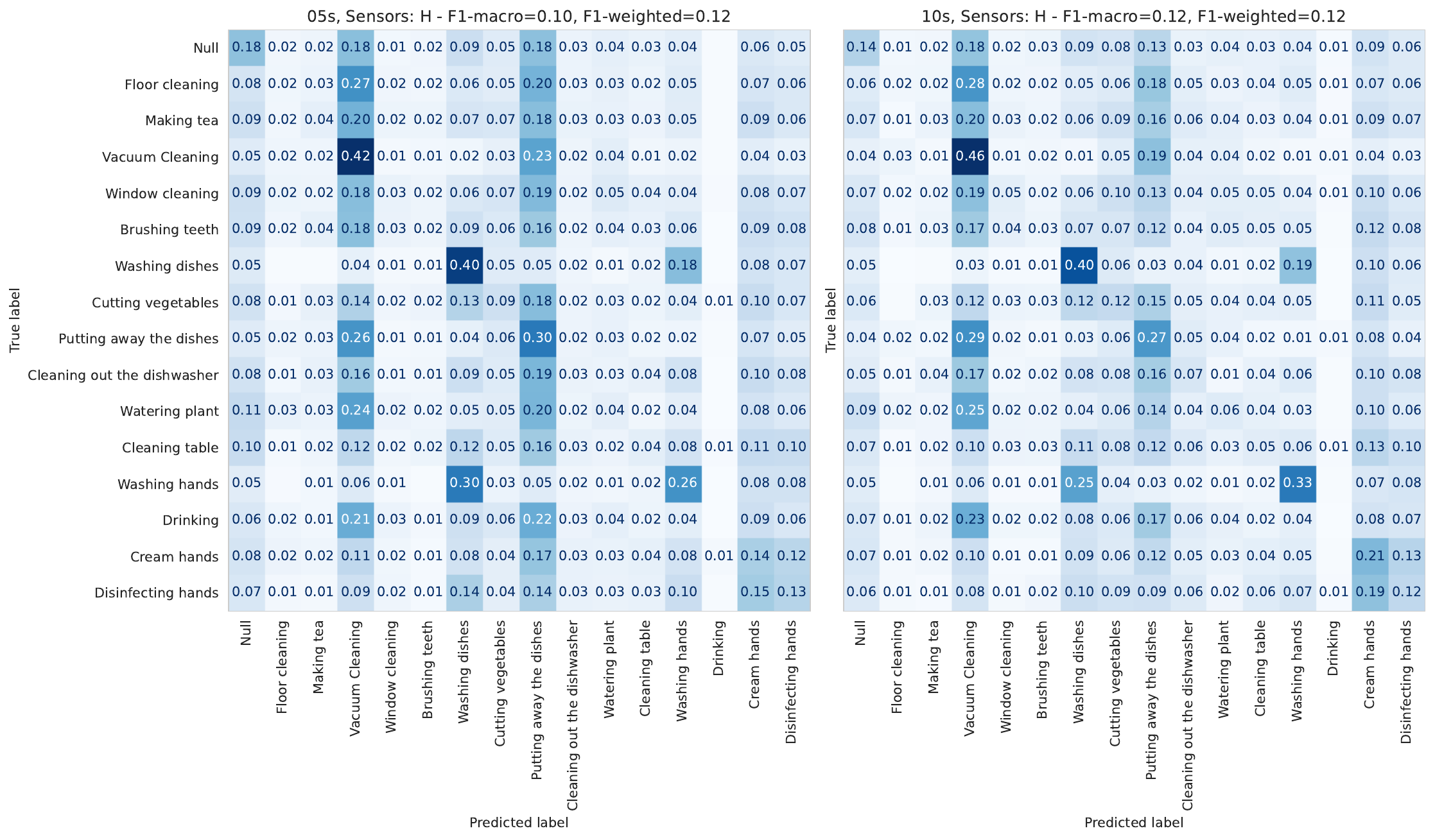}
\caption{Confusion matrices for models trained with humidity as the only modality.}
\label{afig:conf_h}
\end{figure}
\begin{figure}
\includegraphics[width=\linewidth]{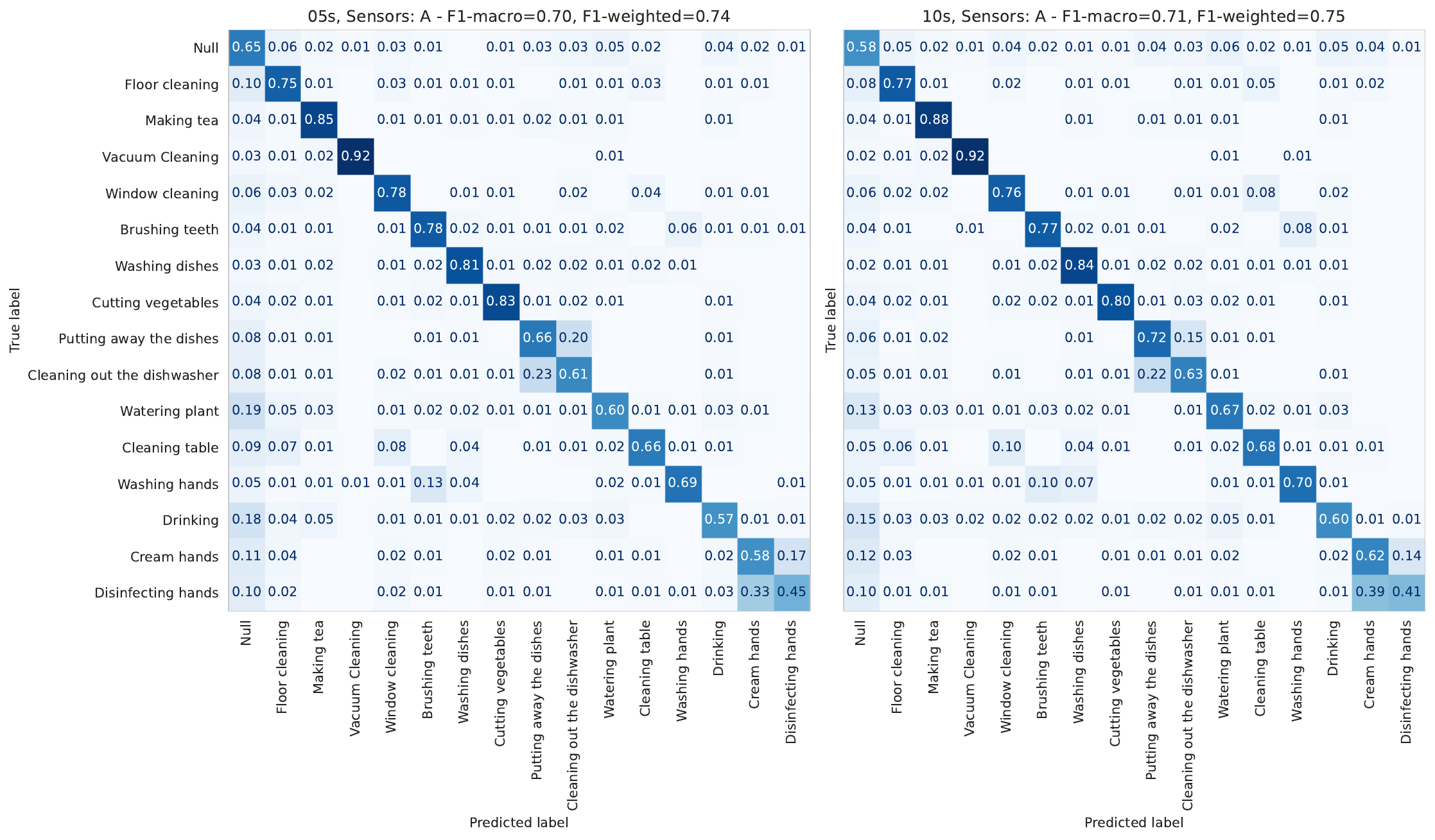}
\end{figure}
\begin{figure}
\includegraphics[width=\linewidth]{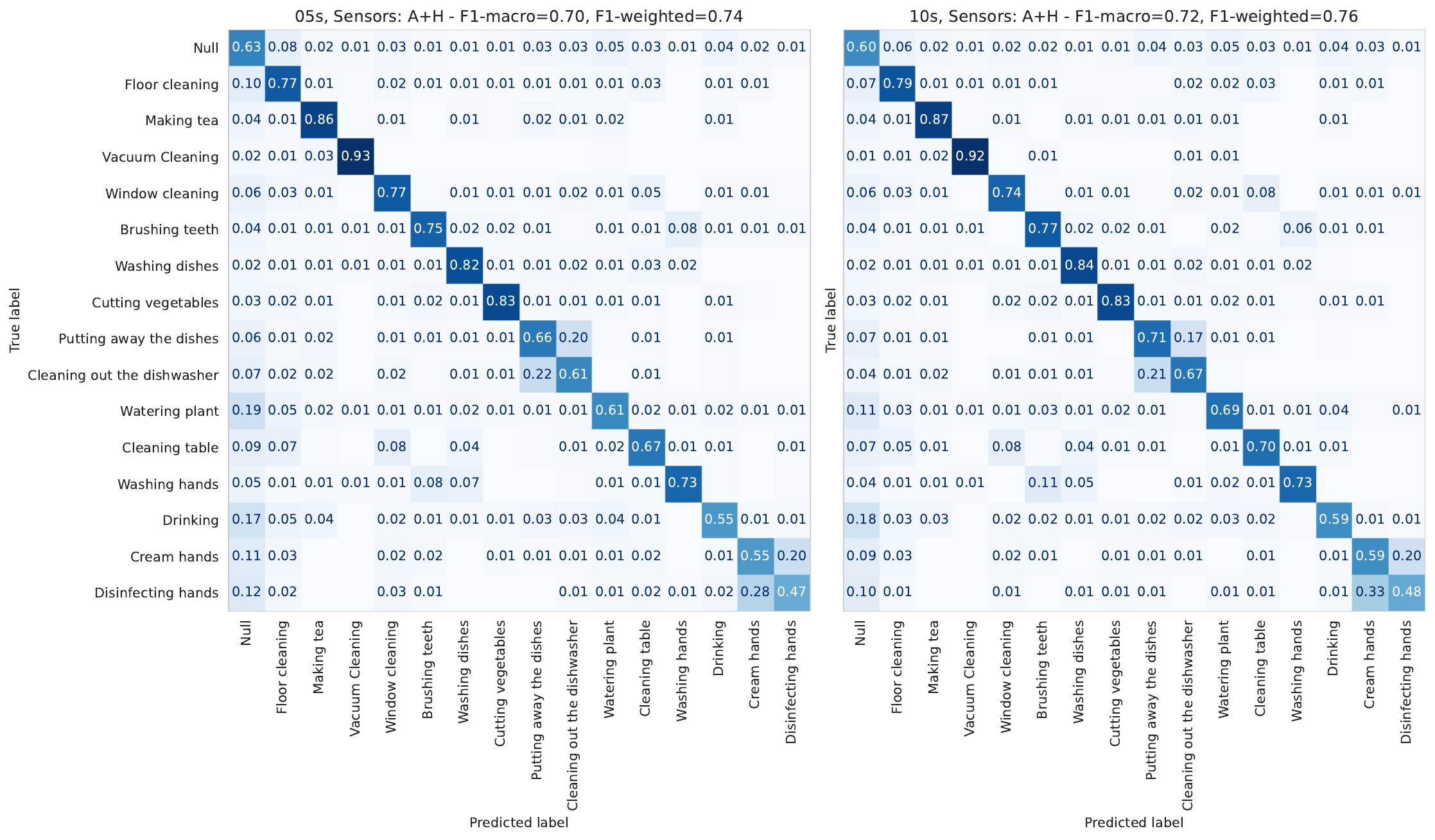}
\end{figure}
\begin{figure}
\includegraphics[width=\linewidth]{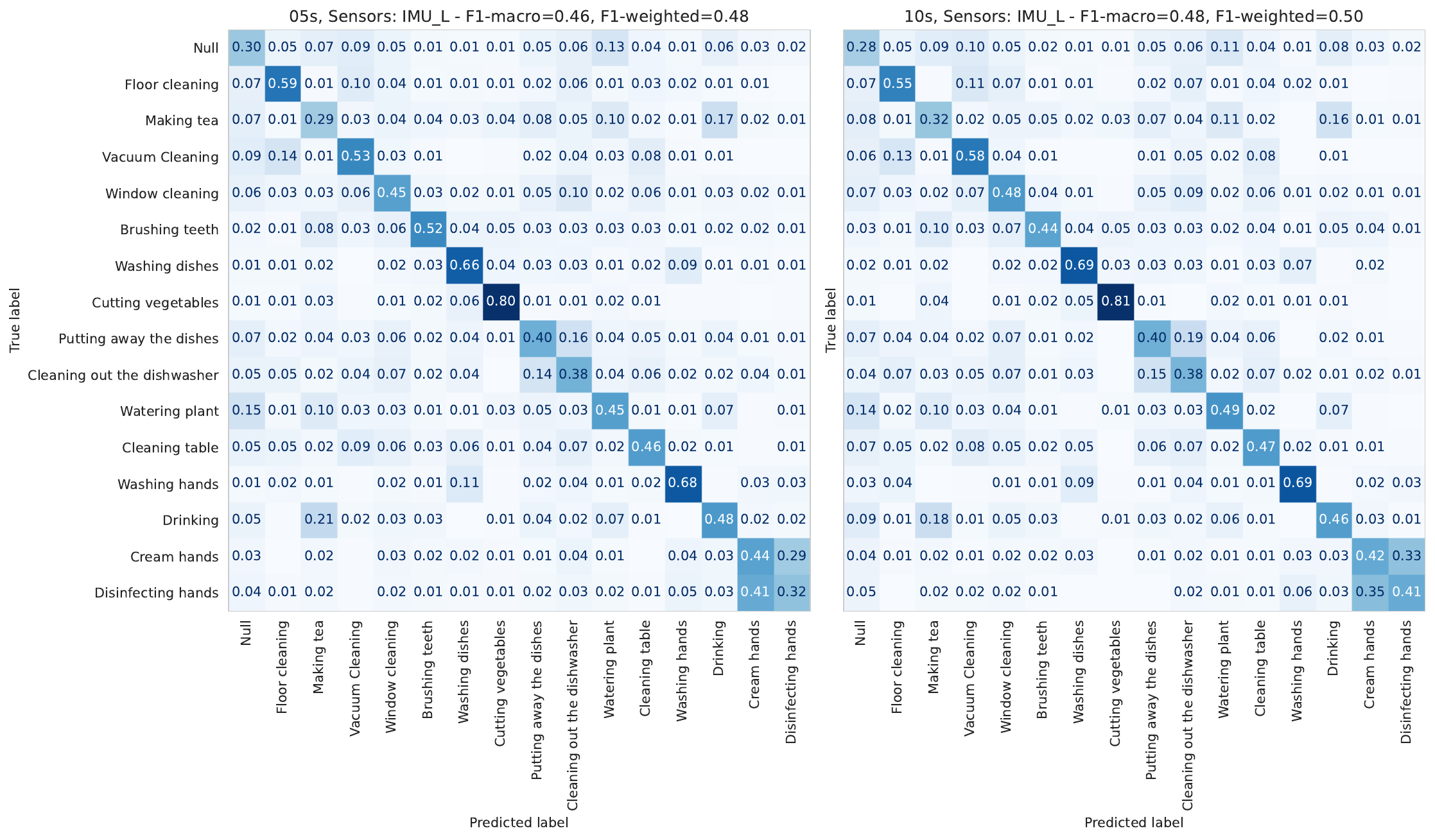}
\end{figure}
\begin{figure}
\includegraphics[width=\linewidth]{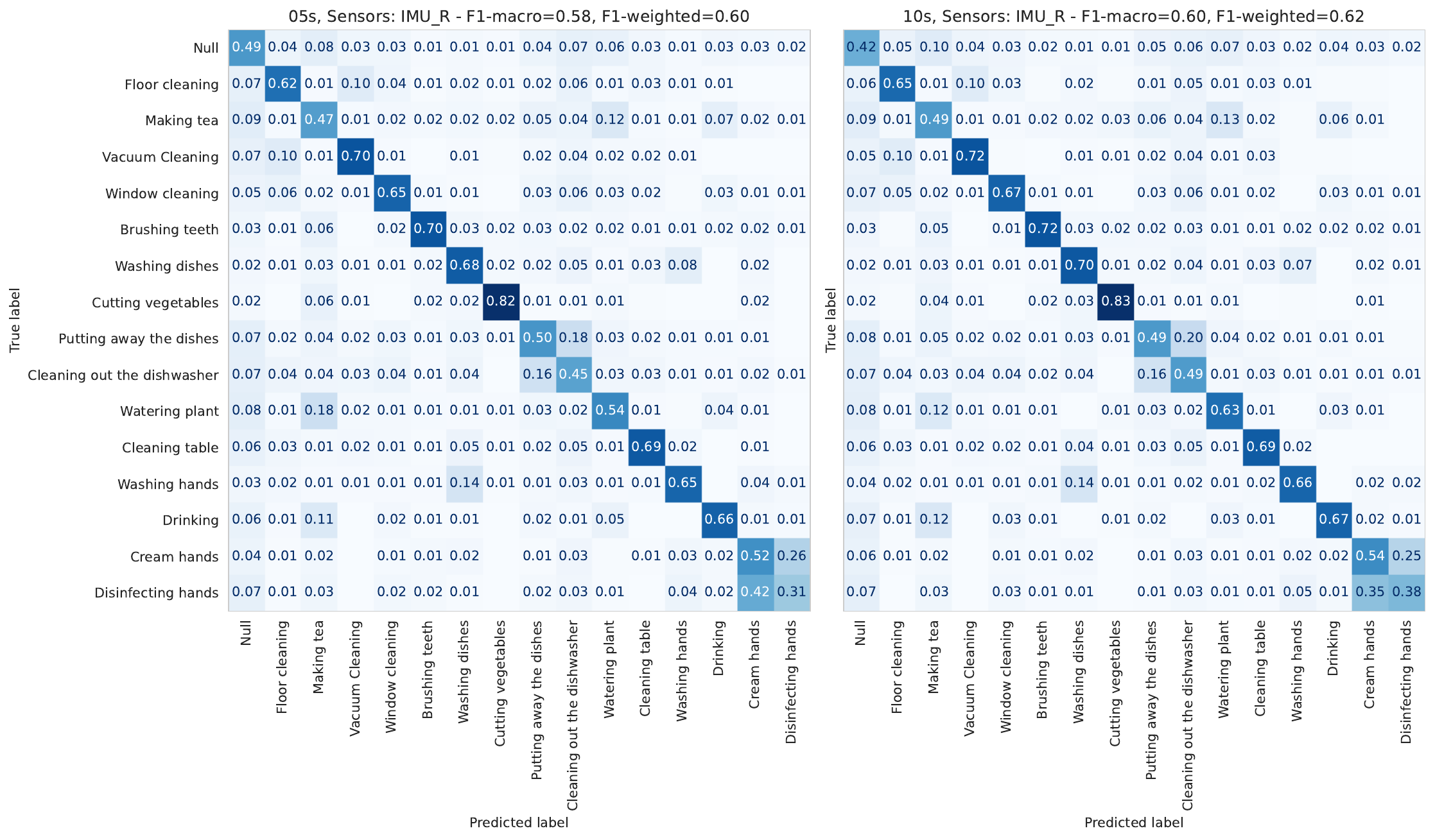}
\end{figure}
\begin{figure}
\includegraphics[width=\linewidth]{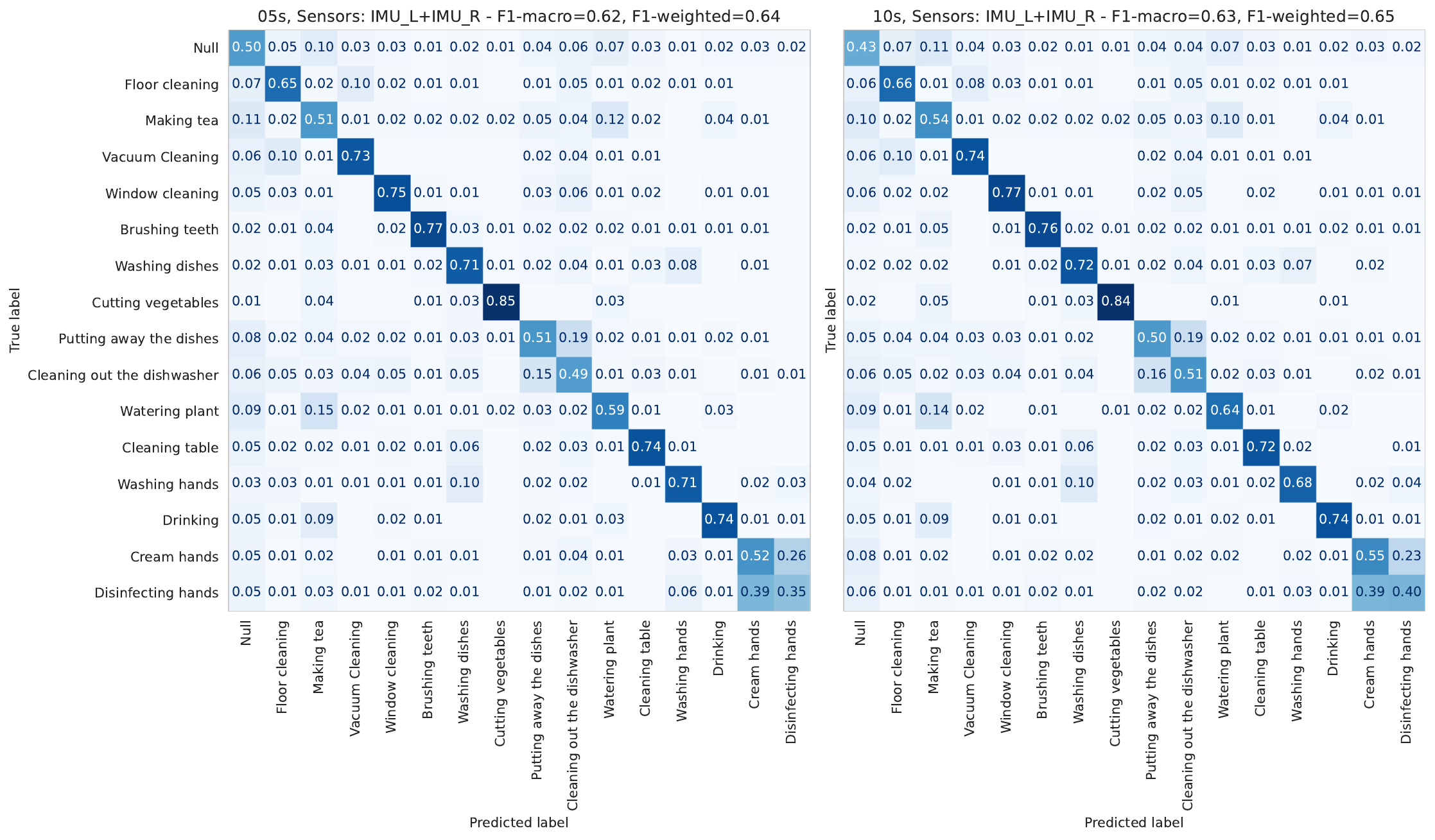}
\end{figure}
\begin{figure}
\includegraphics[width=\linewidth]{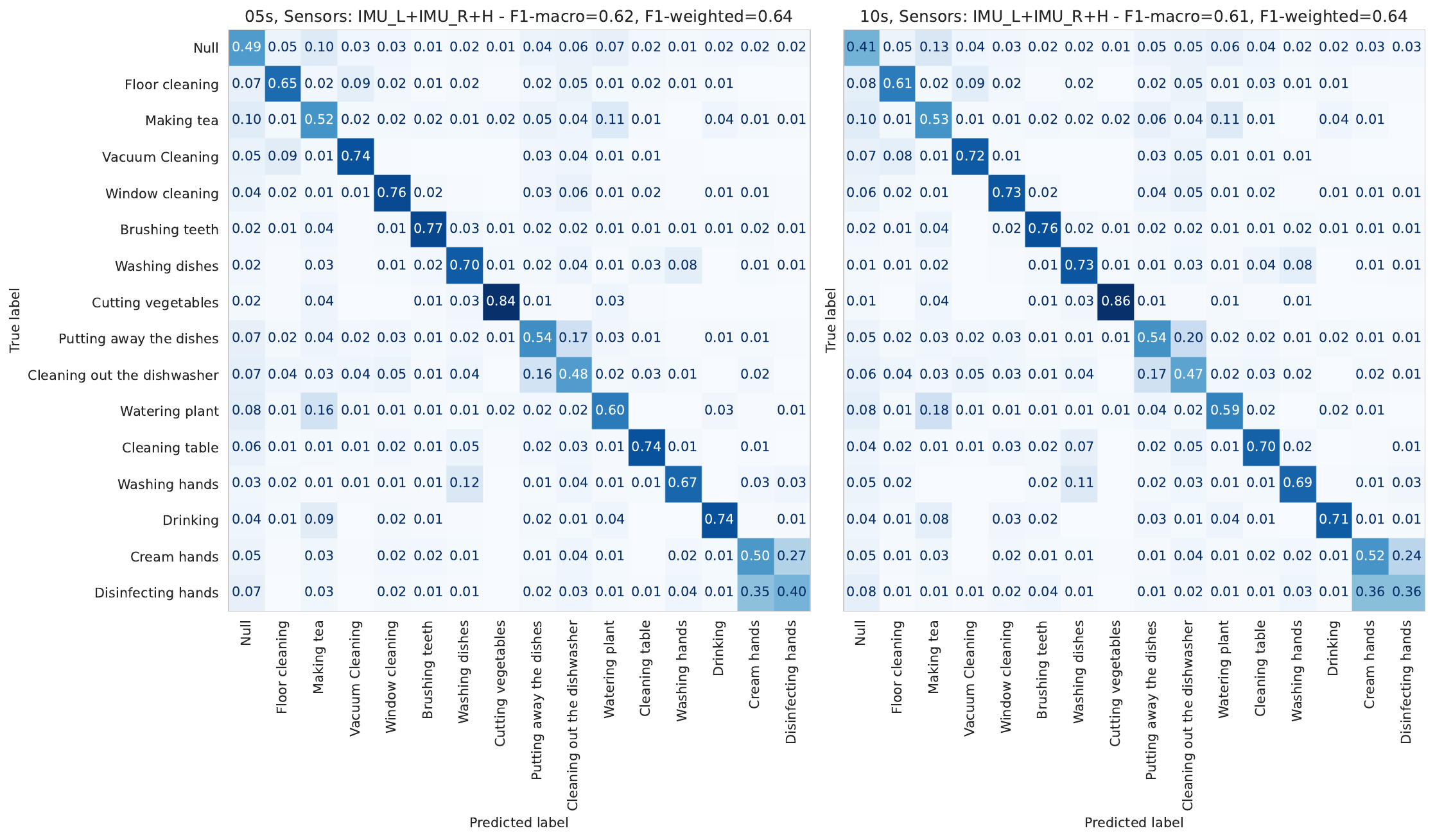}
\end{figure}\begin{figure}
\includegraphics[width=\linewidth]{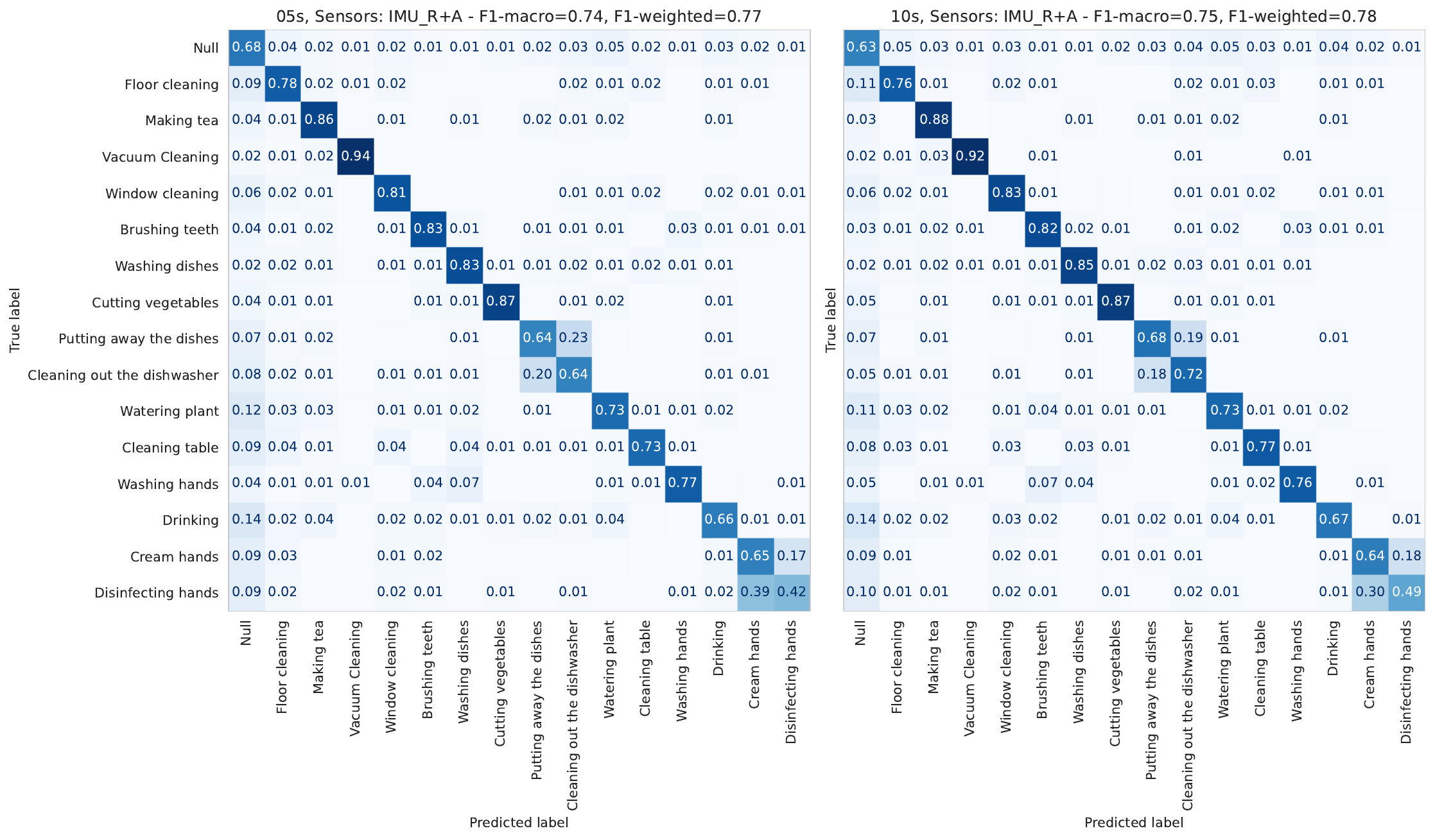}
\end{figure}
\begin{figure}
\includegraphics[width=\linewidth]{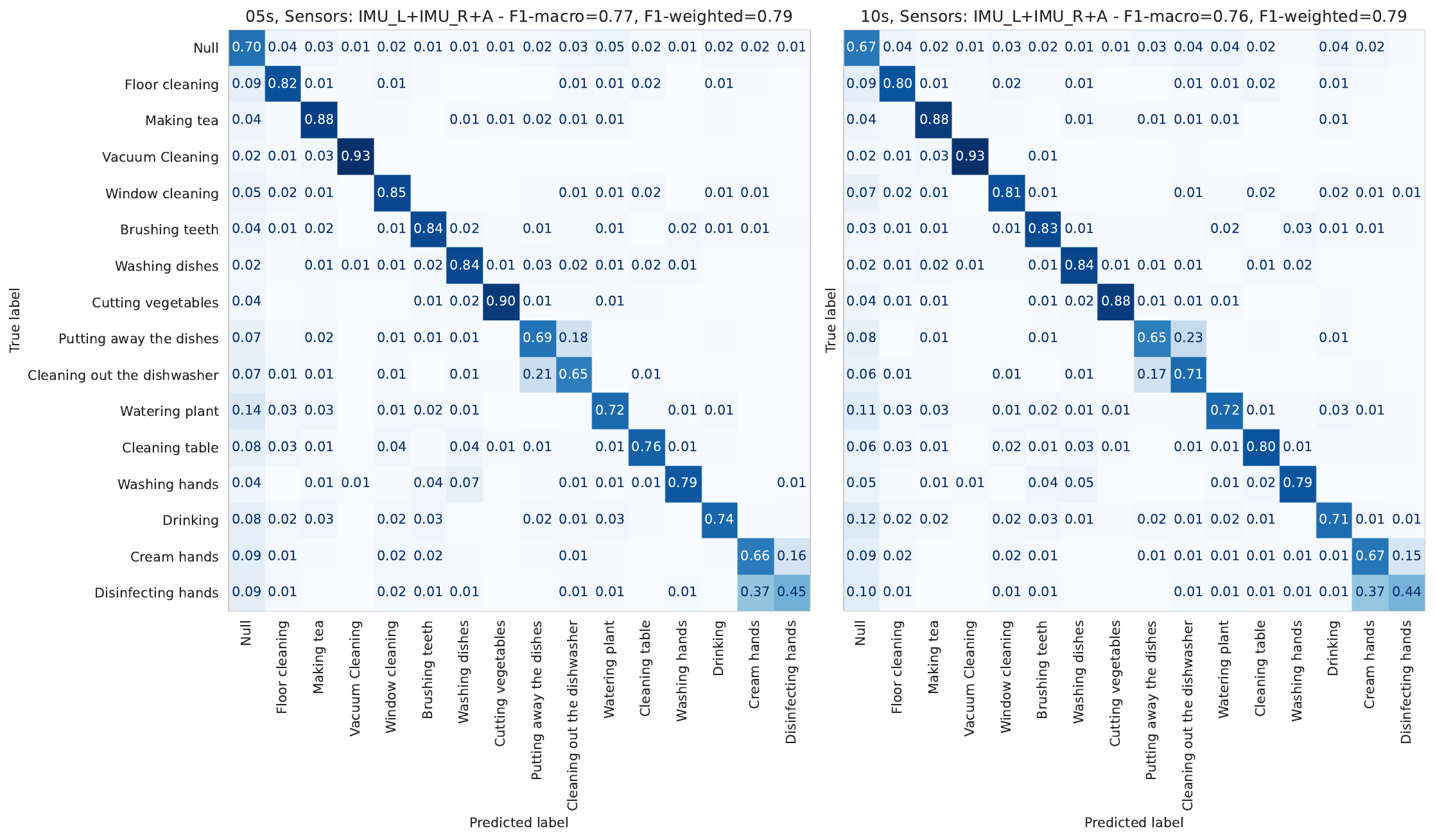}
\end{figure}
\begin{figure}
\includegraphics[width=\linewidth]{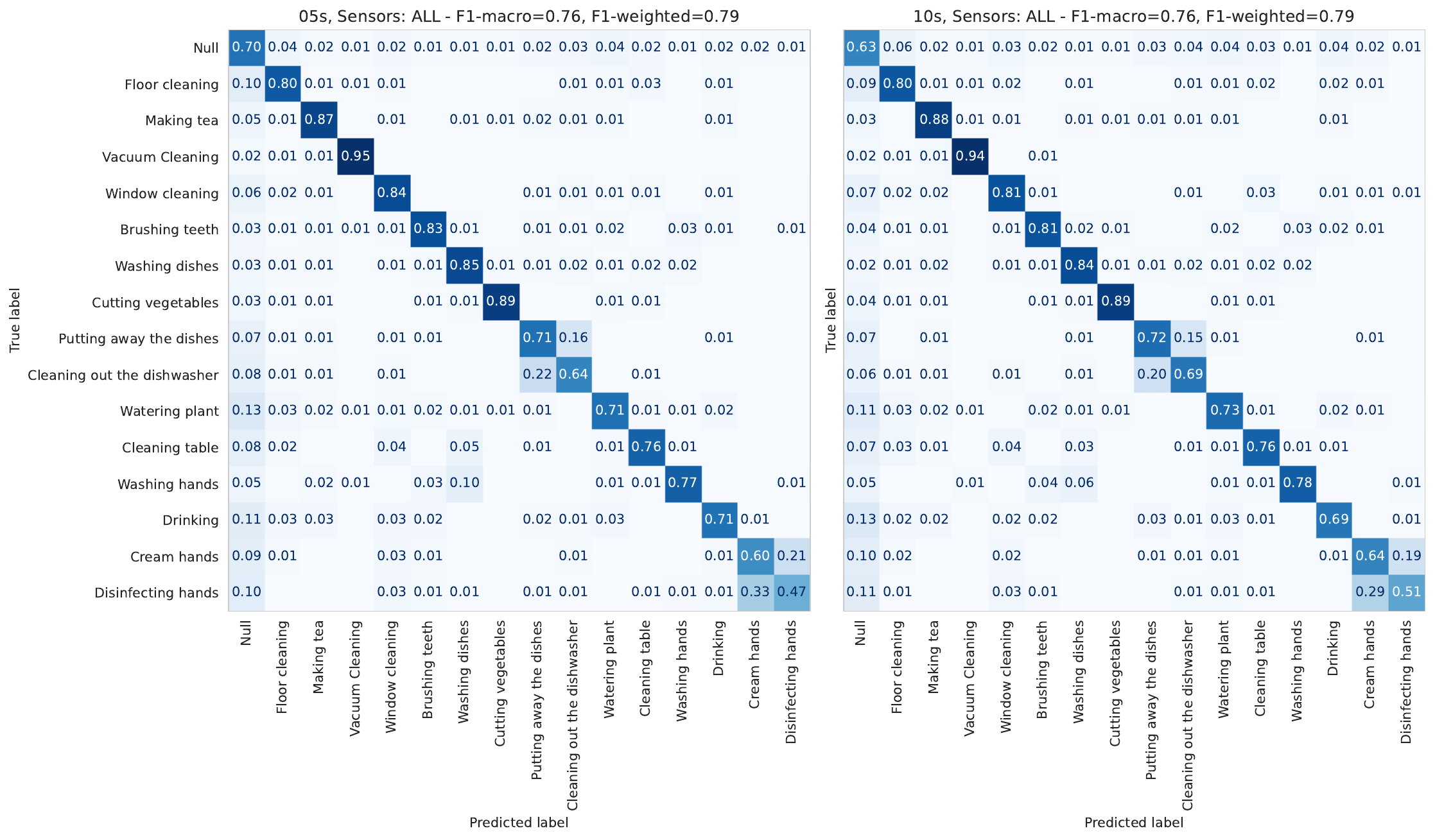}
\end{figure}

\clearpage
\subsection{Per-Participant Results}
In Table \ref{atab:all_results}, we list the complete machine learning results for a window size of 5\,s, showcasing the relatively consistent performance across participants.

\begin{table}
    \centering

    \caption{Per participant machine learning results (5\,s windows) for all sensor combinations. We report the accuracy and the F1-score (macro and weighted).}
    \label{atab:all_results}

\resizebox{\textwidth}{!}{\begin{tabular}{llrrrrrrrrrrrrrrrrrrrr}
\toprule
 & Participant & 01 & 02 & 03 & 04 & 05 & 06 & 07 & 08 & 09 & 10 & 11 & 12 & 13 & 14 & 15 & 16 & 17 & 18 & 19 & 20 \\
Sensors &  &  &  &  &  &  &  &  &  &  &  &  &  &  &  &  &  &  &  &  &  \\
\midrule
\multirow[t]{3}{*}{H} & Accuracy & 0.12 & 0.17 & 0.14 & 0.14 & 0.19 & 0.15 & 0.16 & 0.12 & 0.13 & 0.11 & 0.13 & 0.09 & 0.13 & 0.16 & 0.15 & 0.15 & 0.17 & 0.15 & 0.13 & 0.17 \\
 & F1-Score (macro) & 0.08 & 0.11 & 0.09 & 0.09 & 0.13 & 0.11 & 0.11 & 0.09 & 0.09 & 0.08 & 0.09 & 0.07 & 0.10 & 0.11 & 0.09 & 0.10 & 0.11 & 0.11 & 0.08 & 0.11 \\
 & F1-score (weighted) & 0.09 & 0.13 & 0.10 & 0.11 & 0.16 & 0.12 & 0.12 & 0.11 & 0.12 & 0.10 & 0.13 & 0.08 & 0.12 & 0.14 & 0.12 & 0.12 & 0.13 & 0.12 & 0.10 & 0.13 \\
\cline{1-22}
\multirow[t]{3}{*}{A} & Accuracy & 0.51 & 0.69 & 0.74 & 0.75 & 0.80 & 0.81 & 0.79 & 0.81 & 0.73 & 0.68 & 0.67 & 0.64 & 0.75 & 0.77 & 0.71 & 0.75 & 0.78 & 0.76 & 0.82 & 0.81 \\
 & F1-Score (macro) & 0.50 & 0.64 & 0.69 & 0.72 & 0.77 & 0.78 & 0.76 & 0.77 & 0.67 & 0.62 & 0.64 & 0.60 & 0.72 & 0.73 & 0.68 & 0.71 & 0.75 & 0.72 & 0.79 & 0.74 \\
 & F1-score (weighted) & 0.52 & 0.68 & 0.75 & 0.75 & 0.80 & 0.81 & 0.80 & 0.81 & 0.73 & 0.67 & 0.67 & 0.64 & 0.75 & 0.76 & 0.70 & 0.75 & 0.78 & 0.77 & 0.82 & 0.81 \\
\cline{1-22}
\multirow[t]{3}{*}{A+H} & Accuracy & 0.52 & 0.66 & 0.77 & 0.77 & 0.81 & 0.82 & 0.79 & 0.80 & 0.72 & 0.63 & 0.71 & 0.67 & 0.75 & 0.76 & 0.70 & 0.72 & 0.80 & 0.76 & 0.79 & 0.82 \\
 & F1-Score (macro) & 0.49 & 0.60 & 0.72 & 0.74 & 0.76 & 0.79 & 0.76 & 0.76 & 0.66 & 0.57 & 0.67 & 0.63 & 0.73 & 0.72 & 0.66 & 0.68 & 0.76 & 0.71 & 0.74 & 0.77 \\
 & F1-score (weighted) & 0.52 & 0.64 & 0.77 & 0.77 & 0.80 & 0.82 & 0.79 & 0.80 & 0.71 & 0.62 & 0.71 & 0.66 & 0.75 & 0.76 & 0.70 & 0.72 & 0.80 & 0.76 & 0.79 & 0.82 \\
\cline{1-22}
\multirow[t]{3}{*}{ALL} & Accuracy & 0.61 & 0.76 & 0.81 & 0.80 & 0.88 & 0.84 & 0.81 & 0.82 & 0.76 & 0.71 & 0.75 & 0.75 & 0.80 & 0.78 & 0.77 & 0.76 & 0.81 & 0.80 & 0.88 & 0.86 \\
 & F1-Score (macro) & 0.59 & 0.72 & 0.76 & 0.78 & 0.85 & 0.82 & 0.78 & 0.79 & 0.69 & 0.66 & 0.72 & 0.72 & 0.78 & 0.74 & 0.74 & 0.73 & 0.79 & 0.77 & 0.84 & 0.82 \\
 & F1-score (weighted) & 0.61 & 0.75 & 0.81 & 0.80 & 0.88 & 0.85 & 0.82 & 0.82 & 0.75 & 0.71 & 0.76 & 0.75 & 0.81 & 0.78 & 0.77 & 0.77 & 0.82 & 0.81 & 0.88 & 0.86 \\
\cline{1-22}
\multirow[t]{3}{*}{IMU\_L} & Accuracy & 0.45 & 0.48 & 0.49 & 0.45 & 0.64 & 0.51 & 0.43 & 0.42 & 0.49 & 0.43 & 0.42 & 0.42 & 0.52 & 0.45 & 0.48 & 0.41 & 0.53 & 0.50 & 0.51 & 0.46 \\
 & F1-Score (macro) & 0.41 & 0.45 & 0.49 & 0.45 & 0.59 & 0.48 & 0.42 & 0.42 & 0.44 & 0.41 & 0.41 & 0.41 & 0.51 & 0.42 & 0.46 & 0.40 & 0.52 & 0.48 & 0.49 & 0.44 \\
 & F1-score (weighted) & 0.45 & 0.47 & 0.50 & 0.45 & 0.64 & 0.52 & 0.43 & 0.42 & 0.46 & 0.42 & 0.42 & 0.41 & 0.52 & 0.44 & 0.48 & 0.41 & 0.55 & 0.50 & 0.50 & 0.47 \\
\cline{1-22}
\multirow[t]{3}{*}{IMU\_L+IMU\_R} & Accuracy & 0.59 & 0.66 & 0.68 & 0.65 & 0.77 & 0.65 & 0.54 & 0.60 & 0.64 & 0.45 & 0.58 & 0.63 & 0.69 & 0.56 & 0.64 & 0.63 & 0.71 & 0.69 & 0.70 & 0.73 \\
 & F1-Score (macro) & 0.58 & 0.64 & 0.66 & 0.65 & 0.73 & 0.64 & 0.52 & 0.59 & 0.61 & 0.45 & 0.58 & 0.62 & 0.68 & 0.53 & 0.62 & 0.62 & 0.68 & 0.66 & 0.67 & 0.68 \\
 & F1-score (weighted) & 0.60 & 0.65 & 0.68 & 0.65 & 0.77 & 0.65 & 0.54 & 0.59 & 0.64 & 0.46 & 0.57 & 0.62 & 0.69 & 0.55 & 0.64 & 0.63 & 0.71 & 0.69 & 0.70 & 0.73 \\
\cline{1-22}
\multirow[t]{3}{*}{IMU\_L+IMU\_R+A} & Accuracy & 0.60 & 0.77 & 0.80 & 0.81 & 0.87 & 0.87 & 0.80 & 0.81 & 0.79 & 0.72 & 0.76 & 0.77 & 0.82 & 0.79 & 0.76 & 0.80 & 0.85 & 0.79 & 0.85 & 0.85 \\
 & F1-Score (macro) & 0.61 & 0.72 & 0.76 & 0.80 & 0.83 & 0.84 & 0.77 & 0.78 & 0.74 & 0.68 & 0.73 & 0.76 & 0.79 & 0.75 & 0.73 & 0.77 & 0.83 & 0.75 & 0.81 & 0.81 \\
 & F1-score (weighted) & 0.62 & 0.76 & 0.80 & 0.81 & 0.88 & 0.87 & 0.80 & 0.81 & 0.79 & 0.72 & 0.76 & 0.77 & 0.82 & 0.79 & 0.76 & 0.80 & 0.86 & 0.80 & 0.85 & 0.85 \\
\cline{1-22}
\multirow[t]{3}{*}{IMU\_L+IMU\_R+H} & Accuracy & 0.61 & 0.67 & 0.67 & 0.65 & 0.77 & 0.62 & 0.57 & 0.58 & 0.62 & 0.46 & 0.58 & 0.60 & 0.69 & 0.56 & 0.65 & 0.63 & 0.71 & 0.70 & 0.72 & 0.71 \\
 & F1-Score (macro) & 0.59 & 0.65 & 0.65 & 0.64 & 0.74 & 0.60 & 0.54 & 0.56 & 0.58 & 0.45 & 0.58 & 0.60 & 0.68 & 0.53 & 0.63 & 0.62 & 0.70 & 0.67 & 0.68 & 0.66 \\
 & F1-score (weighted) & 0.62 & 0.67 & 0.67 & 0.65 & 0.77 & 0.61 & 0.56 & 0.58 & 0.62 & 0.47 & 0.58 & 0.59 & 0.69 & 0.56 & 0.66 & 0.64 & 0.71 & 0.71 & 0.72 & 0.71 \\
\cline{1-22}
\multirow[t]{3}{*}{IMU\_R} & Accuracy & 0.58 & 0.62 & 0.64 & 0.64 & 0.76 & 0.64 & 0.48 & 0.58 & 0.61 & 0.32 & 0.55 & 0.61 & 0.69 & 0.40 & 0.58 & 0.62 & 0.69 & 0.63 & 0.67 & 0.71 \\
 & F1-Score (macro) & 0.55 & 0.59 & 0.63 & 0.62 & 0.72 & 0.62 & 0.44 & 0.56 & 0.58 & 0.28 & 0.54 & 0.59 & 0.66 & 0.39 & 0.54 & 0.58 & 0.65 & 0.59 & 0.61 & 0.66 \\
 & F1-score (weighted) & 0.58 & 0.62 & 0.64 & 0.64 & 0.76 & 0.64 & 0.47 & 0.57 & 0.60 & 0.31 & 0.55 & 0.60 & 0.69 & 0.40 & 0.57 & 0.61 & 0.69 & 0.64 & 0.67 & 0.71 \\
\cline{1-22}
\multirow[t]{3}{*}{IMU\_R+A} & Accuracy & 0.54 & 0.73 & 0.80 & 0.80 & 0.86 & 0.85 & 0.78 & 0.80 & 0.77 & 0.67 & 0.70 & 0.71 & 0.80 & 0.76 & 0.75 & 0.79 & 0.84 & 0.79 & 0.85 & 0.85 \\
 & F1-Score (macro) & 0.52 & 0.69 & 0.75 & 0.78 & 0.82 & 0.82 & 0.75 & 0.76 & 0.71 & 0.61 & 0.68 & 0.68 & 0.77 & 0.72 & 0.71 & 0.76 & 0.81 & 0.75 & 0.81 & 0.80 \\
 & F1-score (weighted) & 0.54 & 0.72 & 0.80 & 0.80 & 0.86 & 0.85 & 0.79 & 0.80 & 0.77 & 0.67 & 0.70 & 0.71 & 0.80 & 0.76 & 0.75 & 0.79 & 0.84 & 0.79 & 0.85 & 0.85 \\
\bottomrule
\end{tabular}}

\end{table}

\end{document}